\documentclass[10pt,twocolumn,letterpaper]{article}

\usepackage{cvpr}              

\usepackage{framed,multirow}
\usepackage{amssymb}
\usepackage{latexsym}
\usepackage{url}
\usepackage{graphicx}
\usepackage{amsfonts}
\usepackage{amsmath}
\usepackage[separate-uncertainty=true]{siunitx}
\usepackage{multirow}
\usepackage{svg}
\usepackage{subcaption}
\usepackage[export]{adjustbox}
\usepackage{acro}
\usepackage{placeins}
\usepackage{tablefootnote}
\definecolor{cvprblue}{rgb}{0.21,0.49,0.74}
\usepackage[pagebackref,breaklinks,colorlinks,citecolor=cvprblue,hyperfootnotes=true]{hyperref}
\usepackage[normalem]{ulem}
\usepackage[accsupp]{axessibility} 

\DeclareAcronym{dof}{
  short=DoF,
  long=degree of freedom,
}
\DeclareAcronym{pnp}{
  short=PnP,
  long=perspective-n-point,
}
\DeclareAcronym{psp}{
  short=PSP,
  long=pedicle screw placement,
}
\DeclareAcronym{ir}{
  short=IR,
  long=infrared,
}
\DeclareAcronym{ar}{
  short=AR,
  long=augmented reality,
}
\DeclareAcronym{fps}{
  short=fps,
  long=frames per second,
}
\DeclareAcronym{sdt}{
  short=SDT,
  long=surgical digital twin,
}
\DeclareAcronym{ml}{
  short=ML,
  long=machine learning,
}
\DeclareAcronym{poc}{
  short=PoC,
  long=proof of concept,
}
\DeclareAcronym{or}{
  short=OR,
  long=operating room,
}
\DeclareAcronym{pca}{
  short=PCA,
  long=principal component analysis,
}
\DeclareAcronym{pcd}{
  short=PCD,
  long=point cloud data,
}
\DeclareAcronym{icp}{
  short=ICP,
  long=iterative closest point,
}
\DeclareAcronym{sds}{
  short=SDS,
  long=surgical data science,
}
\DeclareAcronym{vr}{
  short=VR,
  long=virtual reality,
}
\DeclareAcronym{ct}{
  short=CT,
  long=computed tomography,
}
\DeclareAcronym{mri}{
  short=MRI,
  long=magnetic resonance imaging,
}
\DeclareAcronym{cd}{
  short=CD,
  long=chamfer distance,
}
\DeclareAcronym{psnr}{
  short=PSNR,
  long=peak signal-to-noise ratio,
}
\DeclareAcronym{ssim}{
  short=SSIM,
  long=structural similarity index measure,
}
\DeclareAcronym{lpips}{
  short=LPIPS,
  long=learned perceptual image patch similarity,
}
\DeclareAcronym{imu}{
  short=IMU,
  long=inertial measurement unit,
}

\def\paperID{27} 
\def\confName{DCA in MI Workshop @ CVPR}
\def\confYear{2024}

\title{Creating a Digital Twin of Spinal Surgery: A Proof of Concept}

\author{Jonas Hein$^{1,2,}$\footnotemark[1]
\and
Frédéric Giraud$^1$
\and
Lilian Calvet$^1$
\and
Alexander Schwarz$^2$
\and
Nicola Alessandro Cavalcanti$^1$
\and
Sergey Prokudin$^2$
\and
Mazda Farshad$^1$
\and
Siyu Tang$^2$
\and
Marc Pollefeys$^2$
\and
Fabio Carrillo$^1$
\and
Philipp Fürnstahl$^1$
\and
  \\$^1$Balgrist University Hospital, University of Zurich, Zurich, Switzerland\\
  $^2$ETH Zurich, Zurich, Switzerland
}

\begin{document}
\twocolumn[{%
\renewcommand\twocolumn[1][]{#1}%
\maketitle
\begin{center}
    \centering
    \captionsetup{type=figure}
    \adjincludegraphics[width=0.489\linewidth, height=6cm, trim={{0.2\width} {0.25\height} {0.2\width} {0.30\height}}, clip, keepaspectratio]{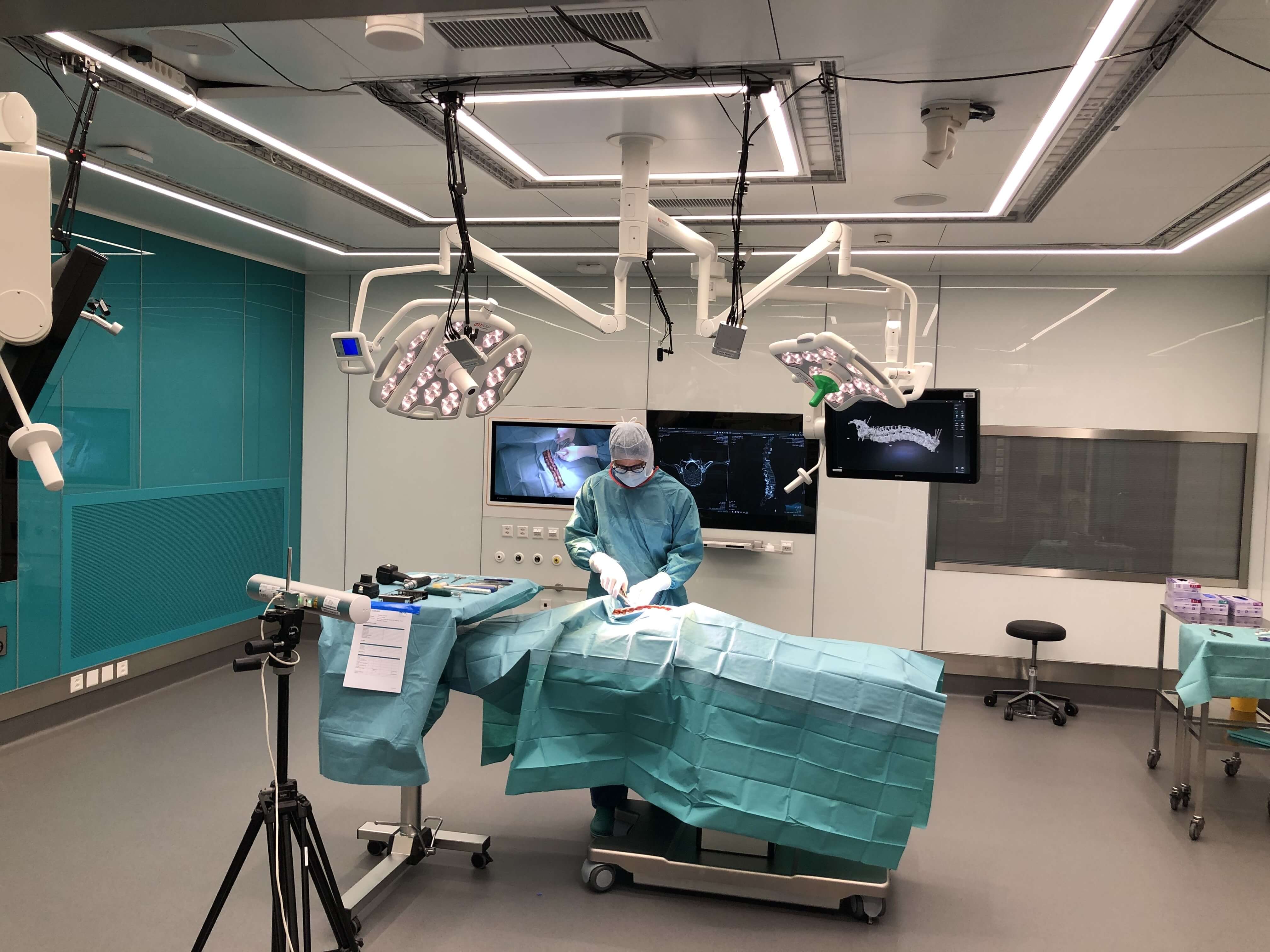}\quad
    \adjincludegraphics[width=0.489\linewidth, height=6cm, trim={{0.2\width} {0.25\height} {0.2\width} {0.30\height}}, clip, keepaspectratio]{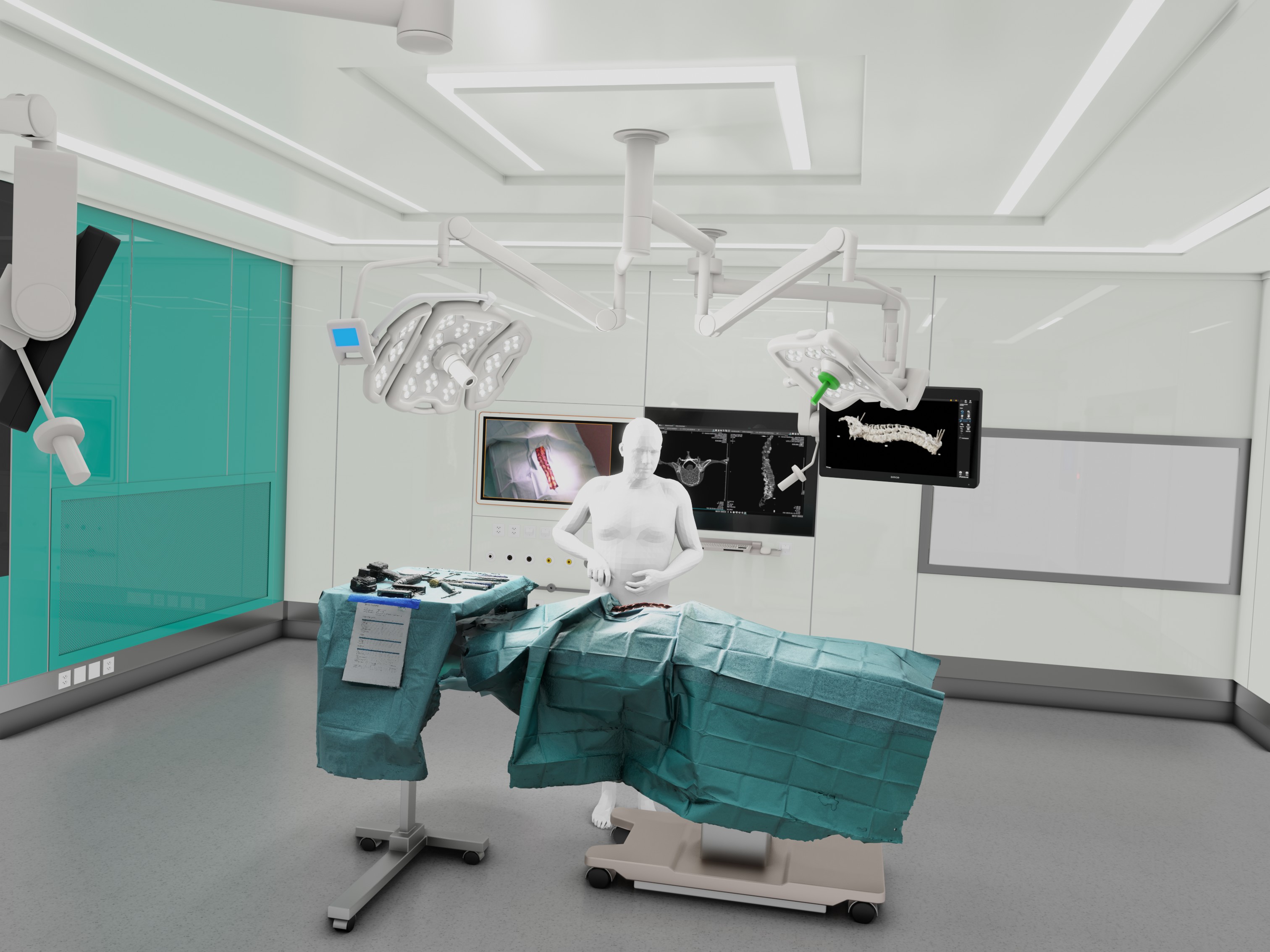}%
    \captionof{figure}{Digital photograph of a spinal surgery (left) and rendering of its digital twin (right) obtained using our proof of concept for surgery digitalization.}
\end{center}%
}]

\begin{abstract} 
\noindent Surgery digitalization is the process of creating a virtual replica of real-world surgery, also referred to as a \ac{sdt}.
It has significant applications in various fields such as education and training, surgical planning, and automation of surgical tasks. 
In addition, \ac{sdt}s are an ideal foundation for machine learning methods, enabling the automatic generation of training data.
In this paper, we present a \ac{poc} for surgery digitalization that is applied to an ex-vivo spinal surgery. 
The proposed digitalization focuses on the acquisition and modelling of the geometry and appearance of the entire surgical scene.
We employ five RGB-D cameras for dynamic 3D reconstruction of the surgeon, a high-end camera for 3D reconstruction of the anatomy, an infrared stereo camera for surgical instrument tracking, and a laser scanner for 3D reconstruction of the operating room and data fusion. 
We justify the proposed methodology, discuss the challenges faced and further extensions of our prototype.
While our \ac{poc} partially relies on manual data curation, its high quality and great potential motivate the development of automated methods for the creation of \ac{sdt}s.
\vspace{10pt}

\end{abstract}

\vspace*{-34pt}
\section{Introduction}\label{sec:introduction}
\vspace*{-3.5pt}
{\let\thefootnote\relax\footnote{*{\tt\scriptsize jonas.hein@inf.ethz.ch}}}

\noindent Surgery digitalization is the process of creating a virtual replica of a real-world surgery, most commonly known as a \acf{sdt}.
The digital twin concept was first introduced by \cite{grieves2014digital} and consists of three main components: a physical object or process along with its environment, its digital replica, and the data and communication links that connect the physical and digital entities. 
It is a specific application of digitalization, namely the process of converting information from a physical format to a digital one, focusing on the replication and simulation of physical entities.
One of the main objectives of a \ac{sdt} is the high-fidelity representation of relevant entities and their interactions during the surgery, including the patient, surgical instruments and devices, as well as medical staff.

Surgery digitalization has diverse downstream applications \citep{mccloy2001virtual}, ranging from optimizing education and enhancing the capabilities of surgical services to enabling the training of surgical robots \citep{xu2021surrol}.
In the realm of education, surgery digitalization has the potential to provide medical \vfill \newpage \noindent students and surgeons with realistic and interactive virtual environments to practice surgical techniques and understand human anatomy without the need for real anatomical models, which are often expensive and scarce, limiting the hands-on learning experience \citep{mao_immersive_2021}. 
It may facilitate operative performance assessment, formative feedback and surgical credentialing by avoiding the need for manual review and assessment of surgical videos \citep{MascagniASAMWAR22,RazaVDK22}.
The ability of replaying or streaming a surgery may enable novel use-cases in the areas of quality control and remote surgery \citep{laaki2019prototyping}.
In the context of workflow optimization, surgery digitization enables the optimization of resource allocations through \ac{ml}-based surgical phase recognition \citep{holm2023dynamic}, or the automatic generation of surgery reports \citep{lin2023sgt++}.
In surgical navigation and robotic surgery, \ac{sdt}s can help to reduce the sim-to-real gap by providing accurate and realistic environments in which robots and \ac{ml}-based applications can be trained before being deployed in the real world \citep{Gumbs21,Barnoyetal21,GumbsGBCSFIHPE22,JecklinJFFE22}.

Surgery digitalization necessitates the fusion of available information from sensors and prior knowledge into a common spatio-temporal representation that accurately describes the state of its physical twin \citep{feussner2017surgery}.
Data may come from different modalities, including imaging, sound and text.
To date, imaging technologies are yet predominant. 
For the acquisition of information associated with anatomy, medical imaging technologies such as \ac{ct}, \ac{mri}, ultrasound, fluoroscopy and endoscopy are preferred. 
Optical cameras remain the solution of choice for other components of the surgery, namely for capturing information associated to medical staff, the \ac{or} and its devices, and surgical instruments \citep{Kadkhodamohammadi17,kadkhodamohammadi2017multi}. 
They are widely used due to their ability to capture detailed visual information in a non-invasive manner while being able to record events in real-time.
Once collected, the data is processed to create a model whose complexity varies depending on the type of information to be encoded and on the downstream applications.
A low level representation may consist of a raw multi-view RGB(-D) video \citep{kadkhodamohammadi2017multi} while a high level representation may consist of a detailed semantic and geometric representation of the surgical scene \citep{ozsoy2021multimodal,holm2023dynamic}. 
High level representations could also include advanced behavioral models such as those proposed by \cite{Zhao23}.

In this work, we describe a \acf{poc} to digitize a segment of spine surgery, more specifically the pedicle screw drilling done within the pedicle screw placement procedure, in near-realistic surgery conditions. 
It deviates from a real surgery in that it is performed by a single surgeon without assistance to mitigate occlusions and limit the number of cameras needed.
The surgery is performed ex-vivo on a human specimen in an operating room dedicated to translational research in surgery\footnote{\url{https://www.or-x.ch/}}, enabling an extensive data collection that would be infeasible during real patient treatment.
The specific problem being addressed is how to combine cutting-edge 3D scanning technologies with optimal data fusion and modelling techniques to create a spatio-temporal 3D model of a surgical scene that verifies the following four criteria: it must be (C1) \textit{faithful with respect to geometry}, meaning that the dimensions and spatial relationships in the model should accurately reflect those of the actual surgical setting over time, (C2) \textit{explicit}, (C3) \textit{modular}, which means it is built from smaller, distinct components that represent real-world objects within the surgery, and (C4) \textit{complete}, encompassing the entire surgical scene to provide a full and uninterrupted representation.
Criterion C1 ensures the model supports highly immersive training and education for surgery. Additionally, it enables precise 3D measurements, essential for surgical navigation, planning and quality control.
Criterion C2 guarantees the model is interpretable, allows for measurements using standard metrics, and is compatible with widely used rendering engines.
Parametric representations, especially for the medical staff and instrument's locations should be prioritized.
Criterion C3 allows for the individual manipulation of different components (anatomy, surgeon, surgical instruments, etc.), enabling customizable simulations and object level reasoning in the context of surgical workflow or activity recognition.
Finally, criterion C4 guarantees a holistic representation of the surgery, from which every downstream application can benefit.
However, this requirement necessitates dedicated data acquisition setups, as the data collected during surgeries today is still too sparse to provide such a holistic representation.

In response to these criteria, we contribute a surgery digitalization approach which generates a \ac{sdt} as a set of textured 3D meshes, representing the furnished \ac{or} and the anatomy (static rigid), the surgeon and the surgical drill (dynamic), in a shared spatio-temporal representation.
The choice of 3D scanning technologies being used and the data fusion and modeling processes are described.
The obtained \ac{sdt} is made publicly available.
Although our \ac{poc} partially relies on manual data curation and assumptions that still diverge from actual surgical settings, it is expected to motivate the development of fully automated and functional methods for surgery digitalization under real surgical conditions.


The remainder of this work is organized as follows. 
The state of the art is discussed in Section \ref{sec:sota}. 
The proposed methodology for surgery digitalization is detailed in \mbox{Section \ref{sec:methodology}}.
Section \ref{sec:results} presents a quantitative and qualitative evaluation of the quality of our \ac{sdt}.
Section \ref{sec:discussion} discusses the proposed methodology and its limitations, perspectives and potential applications, before our conclusions are drawn in Section \ref{sec:conclusion}.

\section{Related work}\label{sec:sota}

\noindent\textbf{Surgical data science} 
Over the last two decades the field of \acl{sds} emerged from the need for systematically captured and structured medical data to improve the quality of interventional healthcare \citep{maier2017surgical}.
The importance of this field further increased with the rapid advance of \ac{ml} methods in the last decade.
State-of-the-art deep learning methods typically require large amounts of structured training data, and the lack thereof is one of the major obstacles in the field \citep{maier2022surgical}.
However, patient-related data is still not systematically recorded and stored.
In high-income countries, which benefit from access to advanced healthcare systems and robust IT infrastructures, difficulties arise from navigating regulatory and policy frameworks, as well as from the high complexity of medical data \citep{zhou_chapter_2020, maier2022surgical}.
Relevant information is often distributed across several disconnected systems and in different data modalities which also makes data collection in a standardized and systematic way highly challenging.
The digitalization of surgeries would enable a more standardized and structured data collection process.

\noindent\textbf{Surgical digital twin}
Digital twins aim to be a perfect virtual representation of their physical counterpart, such that observations of the digital twin yield the same information as observations of the physical one.
Later works extended this original definition by \cite{grieves2014digital} to include physical, bio-mechanical, or behavioral models that enable the simulation, prediction of future states, and closed-loop optimization of task-specific objectives \citep{BjellandRSPSHB22}.
In the medical field, a digital twin of the patient has the potential to enable patient-specific optimal treatment \citep{lonsdale2022perioperative}.
Previous works have proposed digital twins for specific anatomies and interventions, including knee arthroscopy and skull base surgery \citep{BjellandRSPSHB22, shu2023twin}.
Most approaches rely on the registration of a preoperative 3D model of the patient-specific anatomy with the patient and the surgical instruments, typically through marker-based tracking.
However - to the best of our knowledge - no existing model aims to capture a full surgery yet.
On the scale of an operating room, the interactions between patient, instruments, surgeons and medical personnel are highly relevant for an accurate description of the current state of the surgery.
Surgical scene graphs \citep{ozsoy2021multimodal} are a lightweight representation of high-level spatial and semantic relationships of entities in the \ac{or}.
Several works proposed to estimate surgical scene graphs from video \citep{ozsoy_labrad-or_2023}.
Similarly, surgical process models \citep{neumuth2017surgical} have been proposed to hierarchically describe the surgical phases and steps comprising an intervention.
While these graph-based representations may be beneficial for high-level tasks such as visual question answering \citep{yuan2023advancing} or surgical phase recognition \citep{holm2023dynamic}, they abstract the low-level geometry required for a high-fidelity representation of the surgical scene.

\begin{figure*}[t]
    \centering
    \includegraphics[width=\linewidth, keepaspectratio]{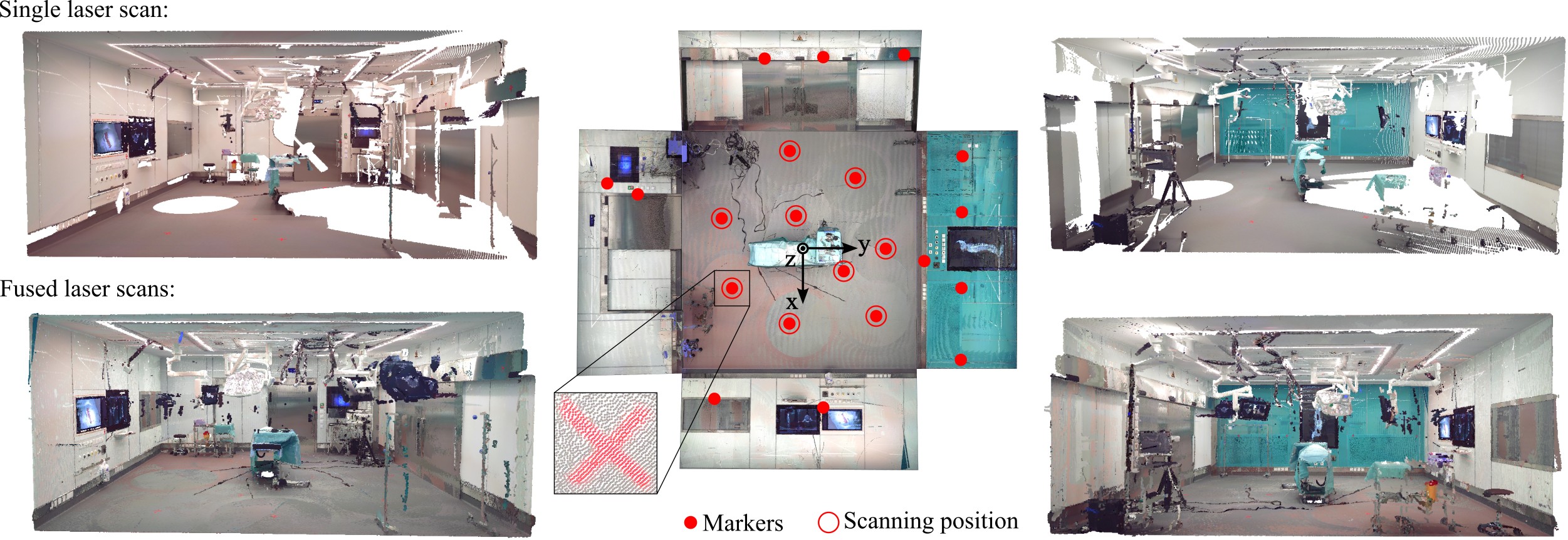}
    \caption{Generation of the reference point cloud from multiple laser scans. The first row shows the point cloud obtained from a single laser scan, illustrating the occlusion challenge. In comparison, the bottom row shows the reference point cloud after fusing all 8 scans. The top view in the center indicates the 21 marker locations and 8 scanning positions within the room. We also indicate the origin of the reference frame, which lies in the ground plane.}
    \label{fig:marker_location_in_or}
\end{figure*}

\noindent\textbf{3D reconstruction} 
Various types of technologies have been developed to digitize the 3D structure of the physical world with high fidelity.
Laser scanning and structured light scanning have established themselves as powerful tools for their direct and precise acquisition of 3D data.
Laser scanning, utilizing the principle of light detection and ranging (LIDAR), offers unparalleled accuracy in capturing large-scale environments and intricate details over vast distances.
On the other hand, structured light scanning, which projects patterned light onto objects and measures deformations through a camera system, excels in capturing high-resolution surface details of smaller objects within controlled environments.
In clinical research, both LIDAR and structured-light approaches have been explored for the 3D reconstruction of soft tissue \citep{maurice2012structured, edgcumbe2015pico, barreto2022aracam, caccianiga2024dense}.
While these methods provide robust solutions for 3D data acquisition, they present limitations in terms of equipment cost, operational complexity, and environmental constraints, which can hinder their applicability in diverse scenarios.

In contrast, photogrammetry, which reconstructs 3D models from 2D optical images, is an alternative that offers versatility and accessibility beyond the capabilities of laser or structured-light scanning.
The literature on computer vision presents photogrammetry methods that are capable of deducing both the shape and appearance of objects from a collection of uncalibrated optical images. 
Recent Neural Radiance Fields (NeRFs) \citep{mildenhall2021nerf}, their extensions to surface reconstruction \citep{wang2023neus2} and large scale acquisitions \citep{li2023neuralangelo}, and even more recently Gaussian splatting techniques \citep{kerbl3Dgaussians}, have demonstrated remarkable performance in both controlled and uncontrolled environments. 
These advances enable the creation of highly detailed and photorealistic renderings from relatively sparse image datasets, marking a significant leap forward in the field's capabilities.
Recent works applying these techniques on medical imaging data have obtained impressive results \citep{gerats2023dynamic, liu2024endogaussian}, however there remain several challenges in their practical application, such as the handling of dynamics and long temporal sequences, or the compatibility with standard rendering engines.


\noindent\textbf{Pose estimation} 
Most of the above-mentioned computer vision methods assume a mainly rigid scene. 
However, a \ac{sdt} setting includes dynamic objects, primarily medical staff, surgical instruments, and the anatomy. 
To estimate the pose of surgical instruments, marker-based navigation systems like the \textit{FusionTrack} (Atracsys LLC, Puidoux, Switzerland), which combine a stereo-camera  with infrared-sensitive markers mounted on instruments, show sub-millimeter accuracy and remain the gold standard solution. 
Their main limitations are the line of sight issue and limited working volume, which have motivated the development of marker-less tracking approaches \citep{doughty2022hmd, hein2023nextgeneration}.
To estimate the pose of the medical staff, motion capture systems can be used.
Vicon systems\footnote{\url{https://www.vicon.com/}} are the gold standard technology for motion capture in film and video game production. 
They are also widely used in sports biomechanics and virtual reality applications.
Multiple high-speed cameras placed around a controlled environment are employed to track reflective markers attached to the subject.
These systems are costly and the required amount of cameras remains invasive in the context of surgery digitalization. 
Other motion capture systems such as the XSens\footnote{\url{https://www.movella.com/products/xsens/}} rely on \acl{imu}s (\acs{imu}s).
However, these systems are impractical for routine captures, largely due to their time-consuming setup and calibration processes.
An easy-to-use alternative are marker-less body pose estimation methods, which have been developed based on computer vision techniques \citep{openpose2019}. 
Similar to marker-based systems, they produce a skeletal representation of the body from image data. 
This skeletal representation seamlessly integrates with parametric models for the human body \citep{SMPL:2015} and hands \citep{MANO:SIGGRAPHASIA:2017}, 
whose parameters optimally explain the image content while facilitating the creation of a surface representation of the body as a dynamic mesh.
Such marker-less pose estimation approaches enable non-invasive data acquisition setups, which is highly relevant for dynamic and restrictive environments such as \ac{or}s.

\section{Methodology}\label{sec:methodology} 

In this section, we describe our prototype for surgery digitalization.
Our data acquisition setup comprises five \mbox{RGB-D} cameras for dynamic 3D reconstruction of the surgeon, a high-end camera for 3D reconstruction of the anatomy, and an \ac{ir} stereo camera for surgical instrument tracking.
We additionally employ a laser scanner for 3D reconstruction of the OR and its devices, and for the fusion of all captured entities in a shared reference frame.
We first describe the acquisition and fusion of data associated with static elements and their modelling in \mbox{Sections \ref{sec:reference_frame}} and \ref{sec:or_tables}, and those associated with dynamic elements in Section \ref{sec:camera_setup}.

\subsection{Reference frame acquisition}\label{sec:reference_frame}

The basis of our \ac{sdt} is a 3D representation of the \ac{or} with metric scale, which serves as our reference frame for the registration of all static and dynamic elements.
We employed a \textit{Faro Focus 3D 120} laser scanner (FARO Technologies Inc., Lake Mary, FL, USA) to generate a point cloud representation of the room.
To minimize occlusions, we conducted 8 scans from various positions, which were subsequently fused. 
To this end, we temporarily and uniformly positioned 21 markers throughout the space, as depicted in Figure \ref{fig:marker_location_in_or}.
These markers were used as point primitives for a point-to-point registration 
of all 8 point clouds. 
Finally, the origin of the reference point cloud was established at the center of the floor. 
Its orientation was defined by the first two main components from principal component analysis applied to the floor points.


In this \ac{poc} we assume that the ceiling objects, \ac{or} equipment, and the instrumentation table are static. 
Based on these assumptions, this reference point cloud is utilized to integrate all components of the model. 

\subsection{Modeling the operating room}
The objective of this phase was to create a detailed and visually accurate virtual model of the operating theatre, including permanently mounted devices like the \ac{or} lamps and displays. 
For this purpose, we utilized the open-source 3D modeling software \textit{Blender} (Stichting Blender Foundation, Amsterdam, Netherlands) in conjunction with its Eevee rendering engine. 
The CAD models of the room and ceiling elements were crafted by a professional graphics artist, while utilizing the reference point cloud to accurately determine the dimensions.
We modeled the textures and materials based on detailed photographs to enhance the visual realism.
Given the necessity for this simulation to accommodate various configurations of the \ac{or}, the ceiling elements in the model were designed with movable joints, which replicate the kinematics of their real-world counterparts. 
Additionally, we incorporated the functionality to adjust the parameters of both the ceiling and \ac{or} lighting, as well as the contents of the display screens within the OR.

The CAD model of the operating theatre is precisely aligned with the reference point cloud. 
In this alignment process, the joints of the ceiling objects are manually adjusted to match the reference point cloud. 
Exemplary renderings can be seen in Figure \ref{fig:operating_room_render}.
Although the manual modeling of the operating room is time-consuming, this one-time effort yields a detailed model of an OR which can be the basis for realistic training simulators and synthetic data generation.

\noindent\textbf{Operating table and anatomy}\label{sec:or_tables}
We employed a photogrammetry approach to reconstruct the operating and instrumentation tables, as well as the visible surface of the anatomy.
For this, we utilized a \textit{Sony Alpha7R} digital single-lens reflex camera (Sony Group Corporation, Tokio, Japan) to capture 102 sets of images from different viewpoints. 
These photos were captured just before the start of the surgery and with the anatomy and instruments already placed on the tables.
Each set included a focus bracket of five pictures to capture finer details of the tools and spine. 
Focus stacking was performed using the publicly available code \citep{focus_stack} applied to the captured photographs.
The commercial photogrammetry software \textit{RealityCapture}\footnote{\url{https://www.capturingreality.com/}} was then used to produce a textured 3D model of the scene from the focus-stacked images.
The use of this software was mainly motivated by the quality of its 3D reconstructions that is competitive with state of the art in computer vision, its good compromise between reconstruction quality and computation time, and its camera self-calibration. 
The focus-stacked images lead to a very fine detailed texture the 3D model benefits from.
Reconstruction artifacts were manually removed.
The feet of both tables were modeled by hand, as their reconstruction was incomplete due to very challenging surface material and occlusions.
The obtained 3D models were then manually aligned with the reference point cloud to accurately reflect the real-world setup.

We furthermore integrated a 3D model of the inner anatomy into our \ac{sdt} as to capture the interactions between the surgical instruments and the anatomy.
To this end, we manually registered the 3D spine model to the visible anatomy surface included in the photogrammetric reconstruction.
Due to time constraints we utilized a generic 3D spine model, however this generic model can be easily replaced with a patient-specific preoperative model.
These models can be obtained from \ac{ct} or \ac{mri} and are readily available for most orthopedic interventions.

\subsection{Motion capture setup} \label{sec:camera_setup}

\begin{figure}
  \centering
  \adjincludegraphics[width=\linewidth, height=6.5cm, trim={{0.0\width} {.1\height} {0.0\width} {0.0\height}}, clip, keepaspectratio]
    {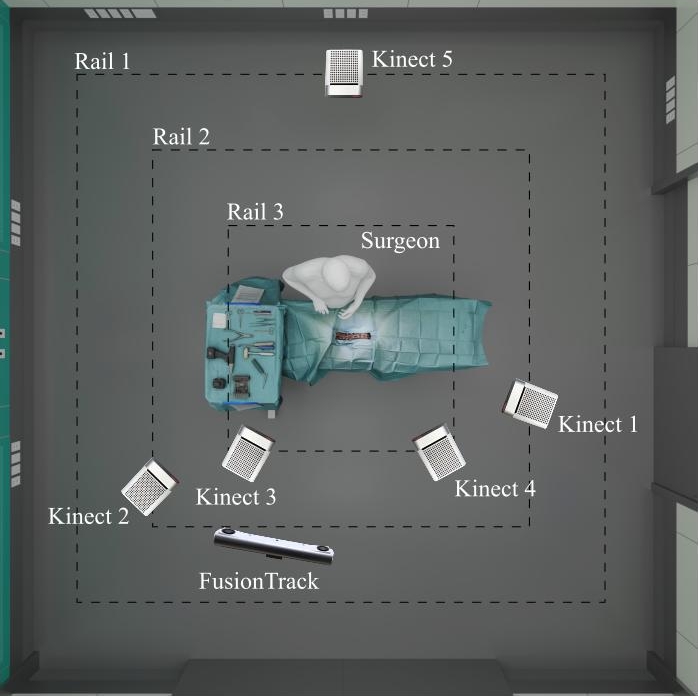}
  \caption{Schematic overview of the experimental setup. Five ceiling-mounted \textit{Azure Kinect} RGB-D cameras capture the motion of the surgeon. A \textit{FusionTrack 500} marker-based tracking system captured the trajectories of the surgical instruments.}
  \label{fig:camera_positions}
\end{figure}

\begin{figure*}[t]
    \centering
    \begin{minipage}[b]{.2\linewidth-1mm}
    \includegraphics[width=\linewidth, keepaspectratio]{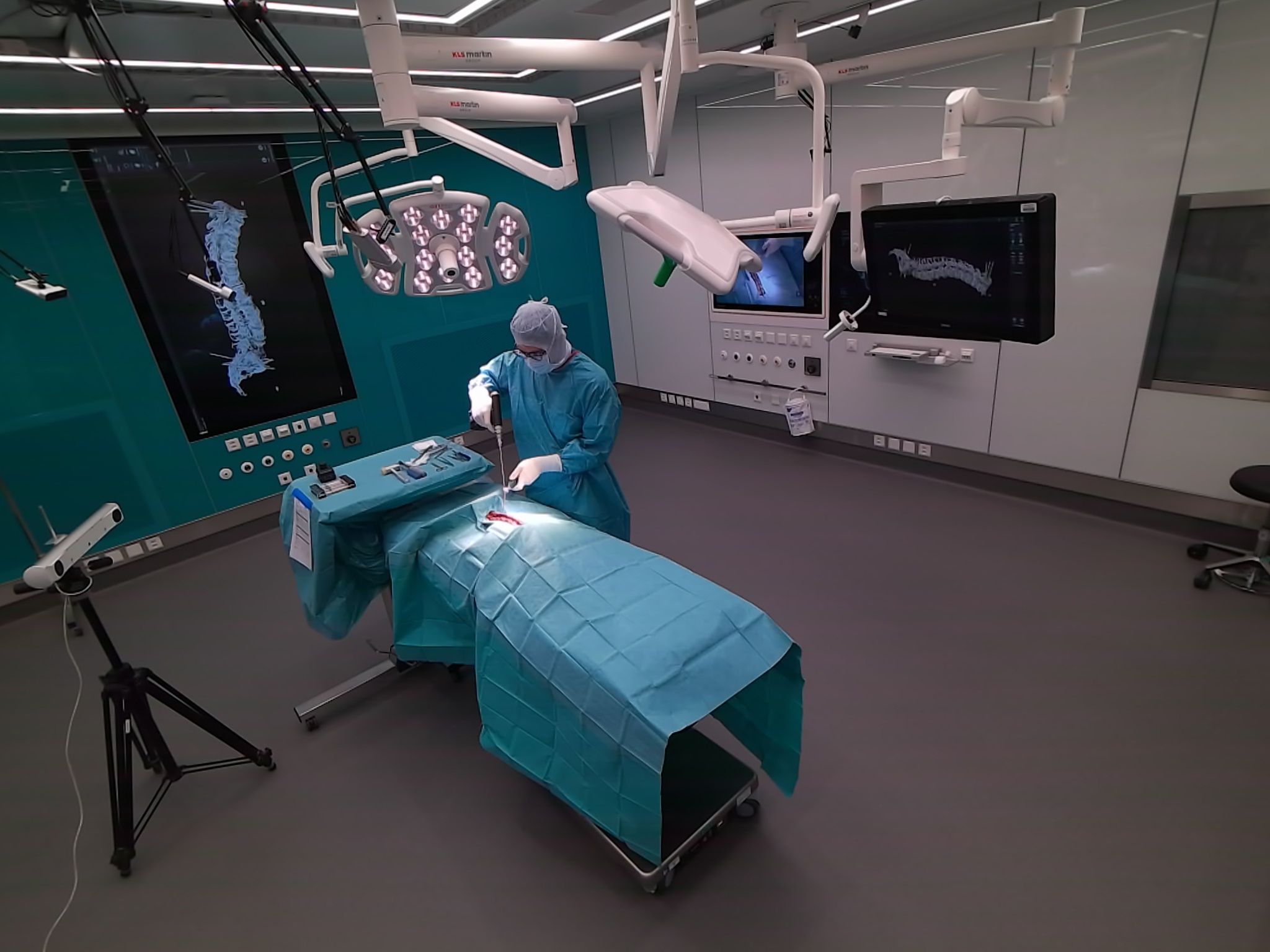}
    \includegraphics[width=\linewidth, keepaspectratio]{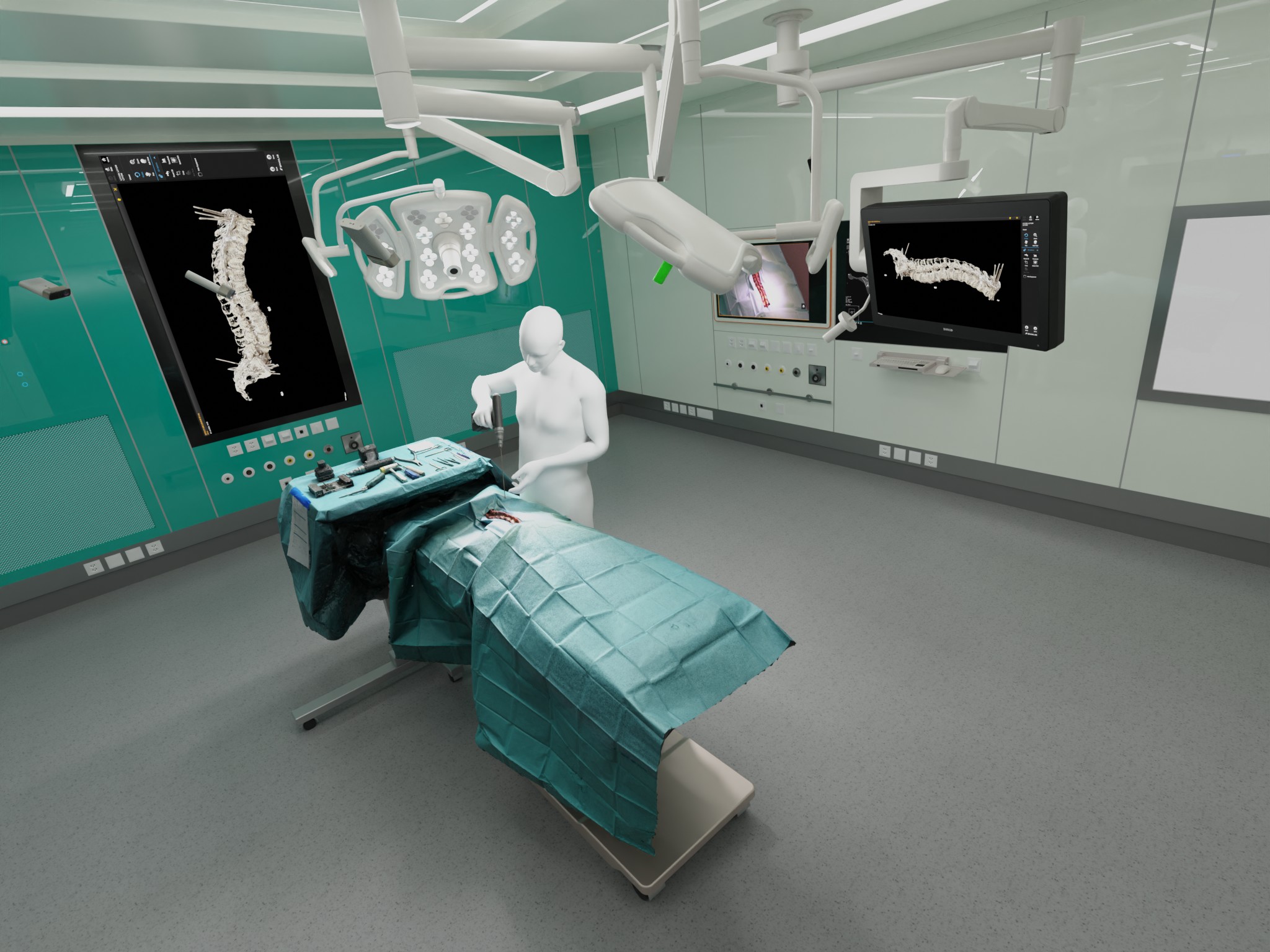}
    \end{minipage}%
    \hspace*{1mm}
    \begin{minipage}[b]{.2\linewidth-1mm}
    \includegraphics[width=\linewidth, keepaspectratio]{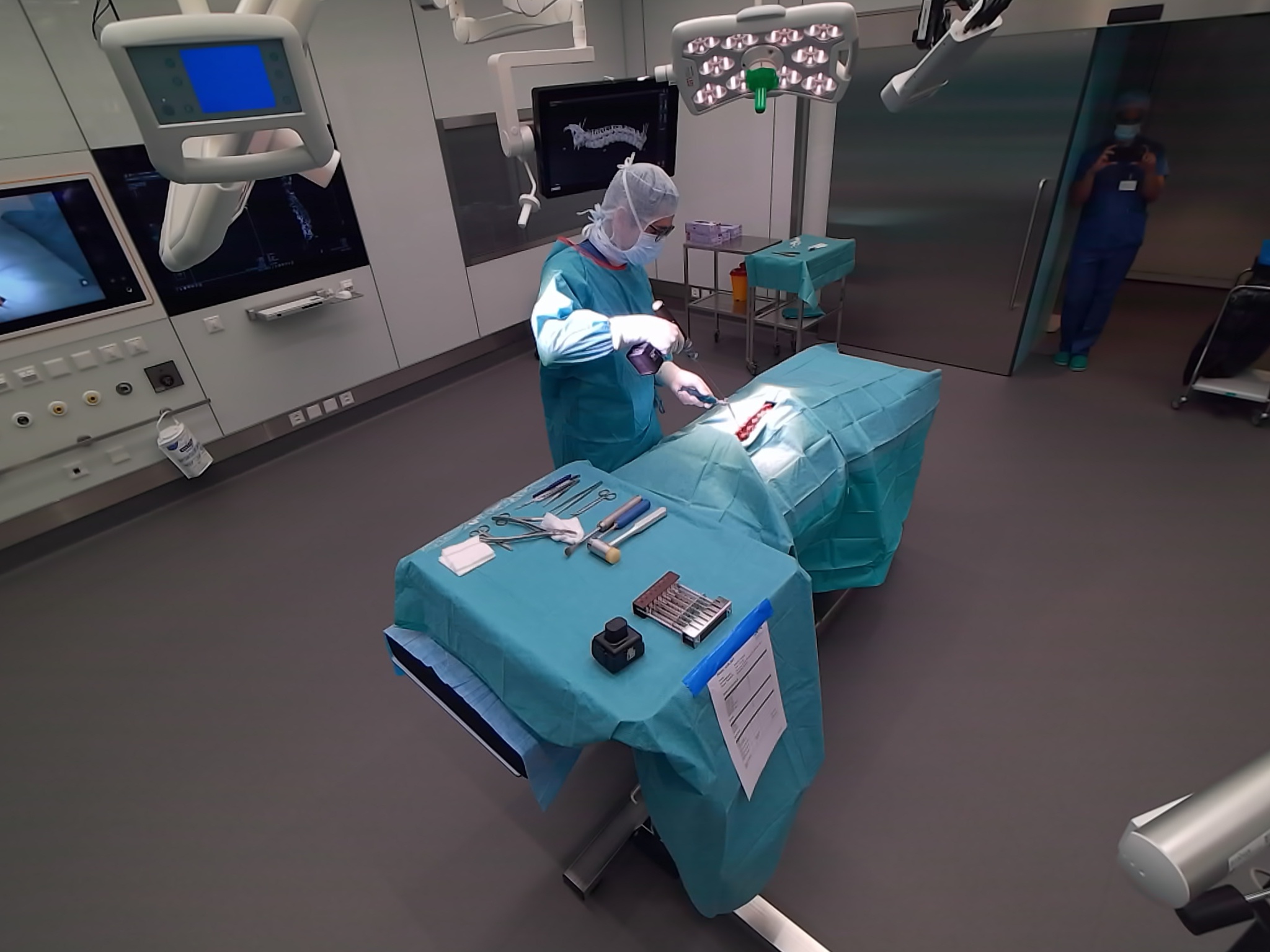}
    \includegraphics[width=\linewidth, keepaspectratio]{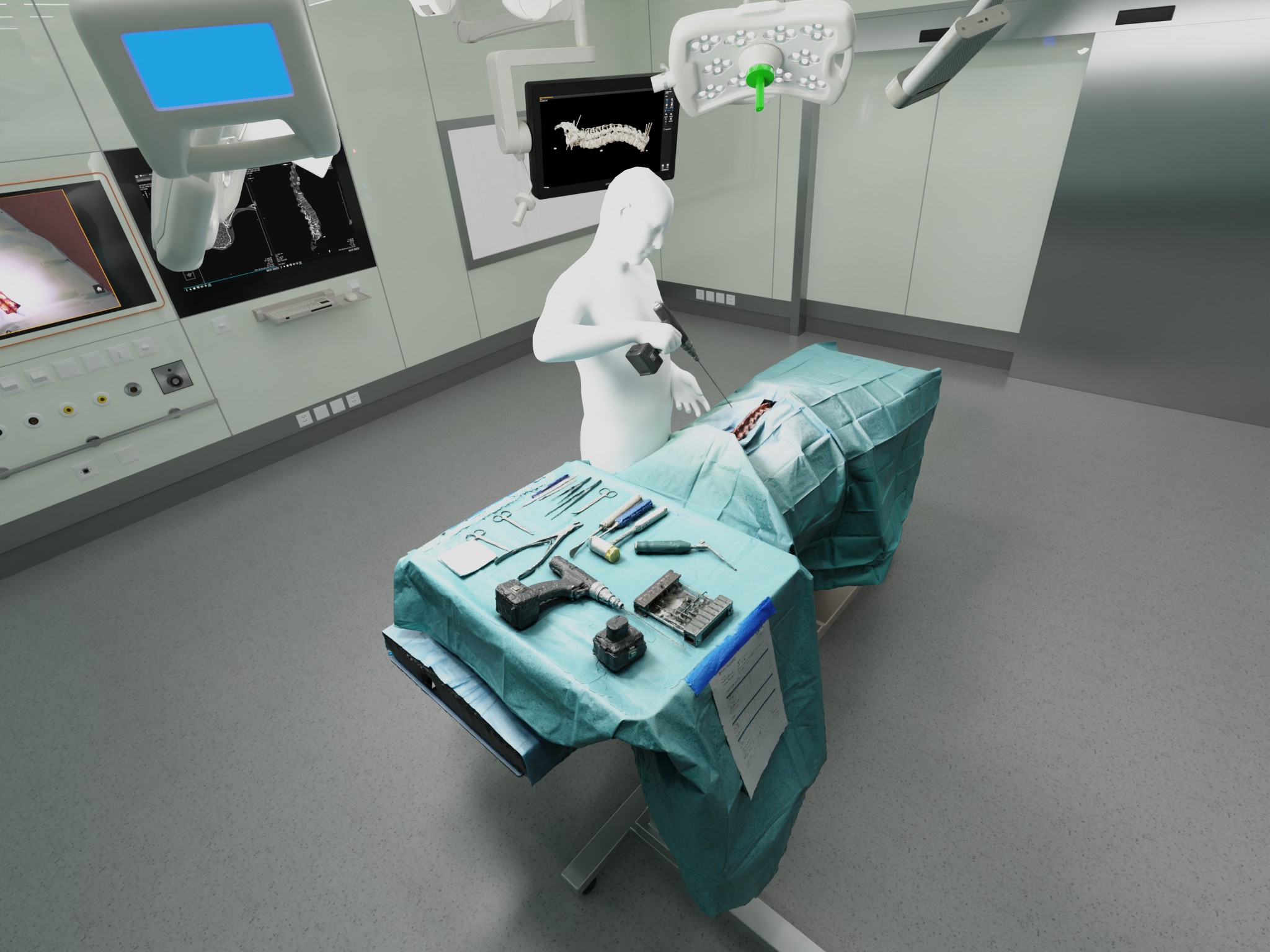}
    \end{minipage}%
    \hspace*{1mm}
    \begin{minipage}[b]{.2\linewidth-1mm}
    \includegraphics[width=\linewidth, keepaspectratio]{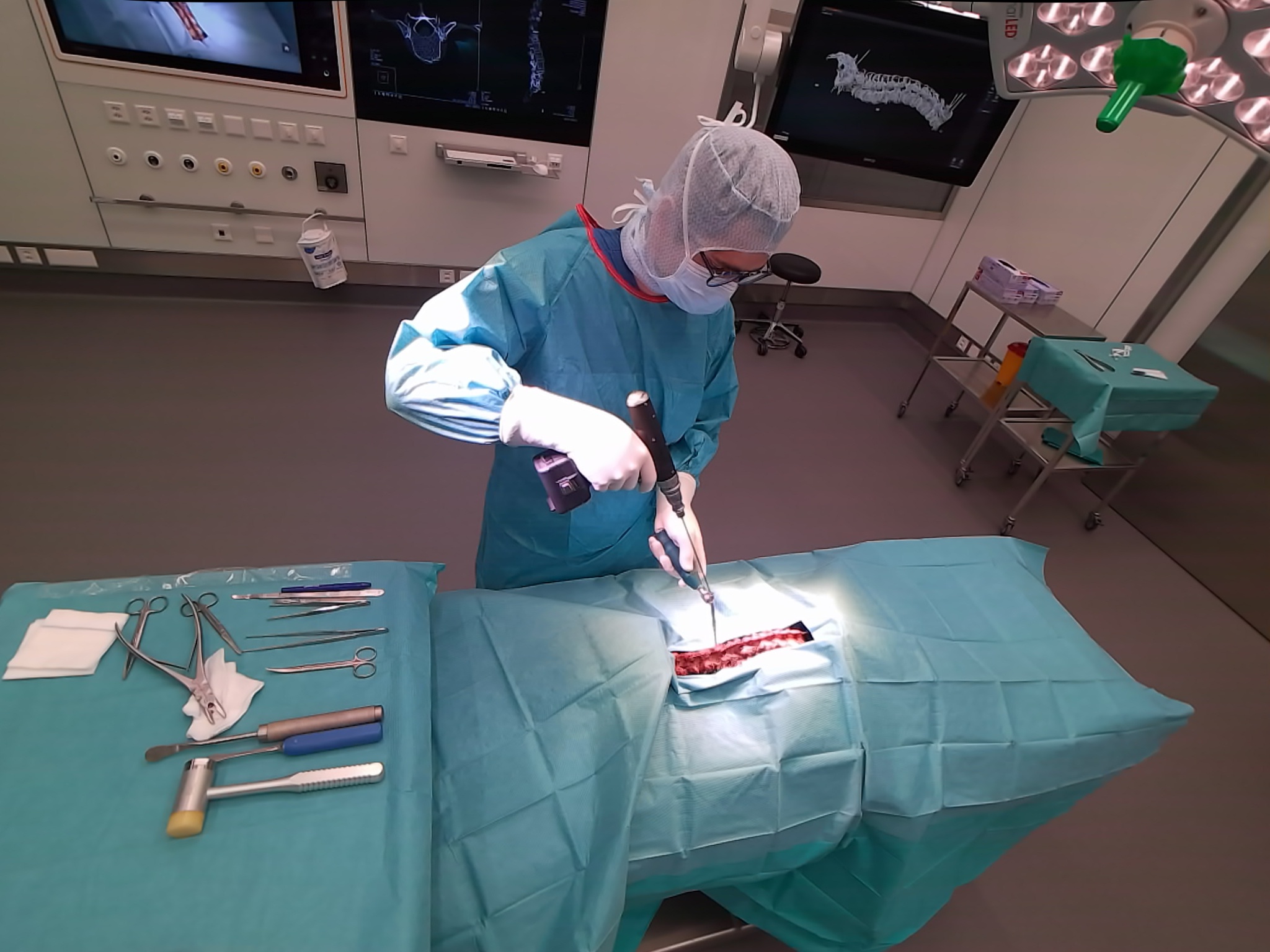}
    \includegraphics[width=\linewidth, keepaspectratio]{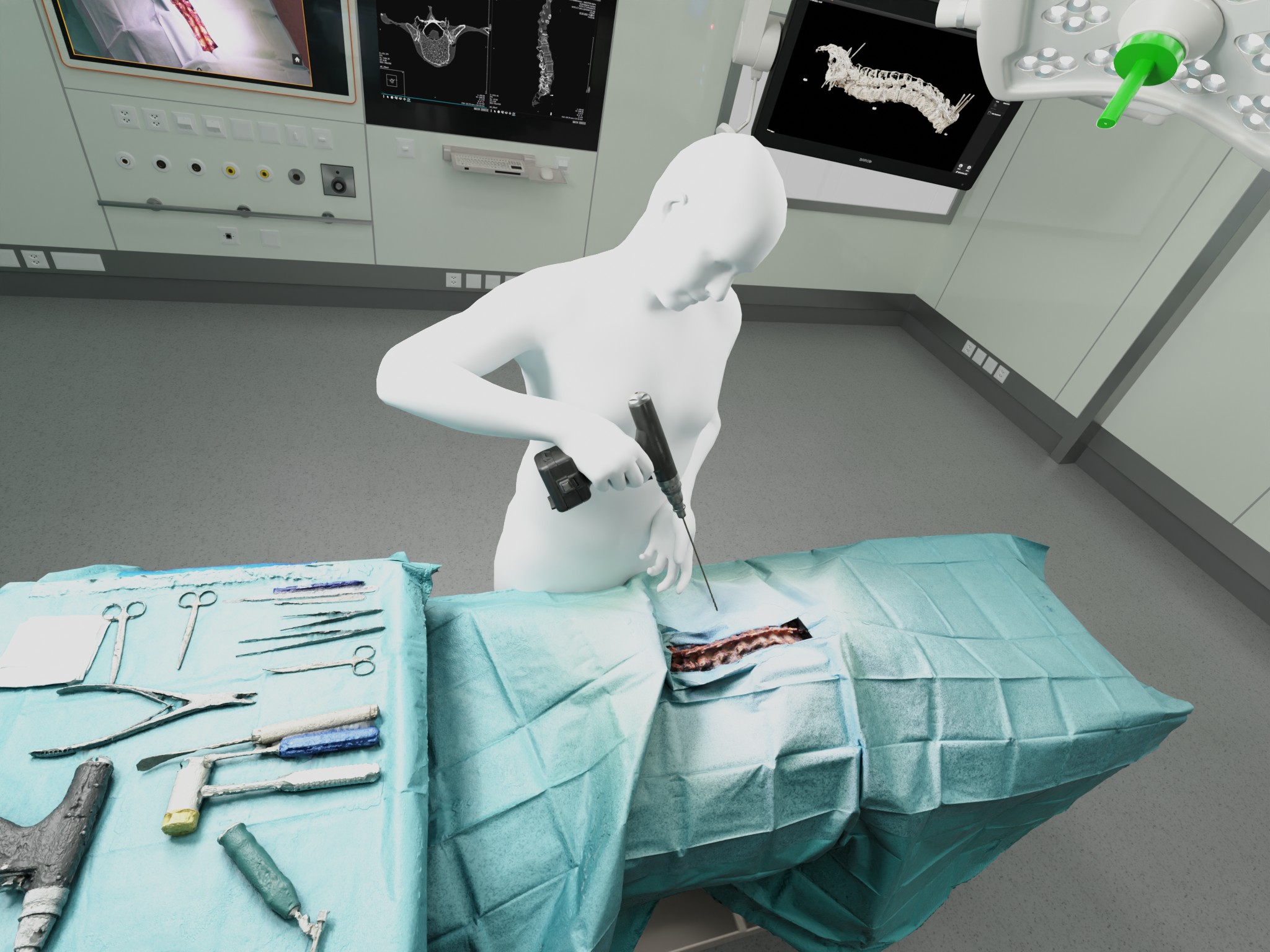}
    \end{minipage}%
    \hspace*{1mm}
    \begin{minipage}[b]{.2\linewidth-1mm}
    \includegraphics[width=\linewidth, keepaspectratio]{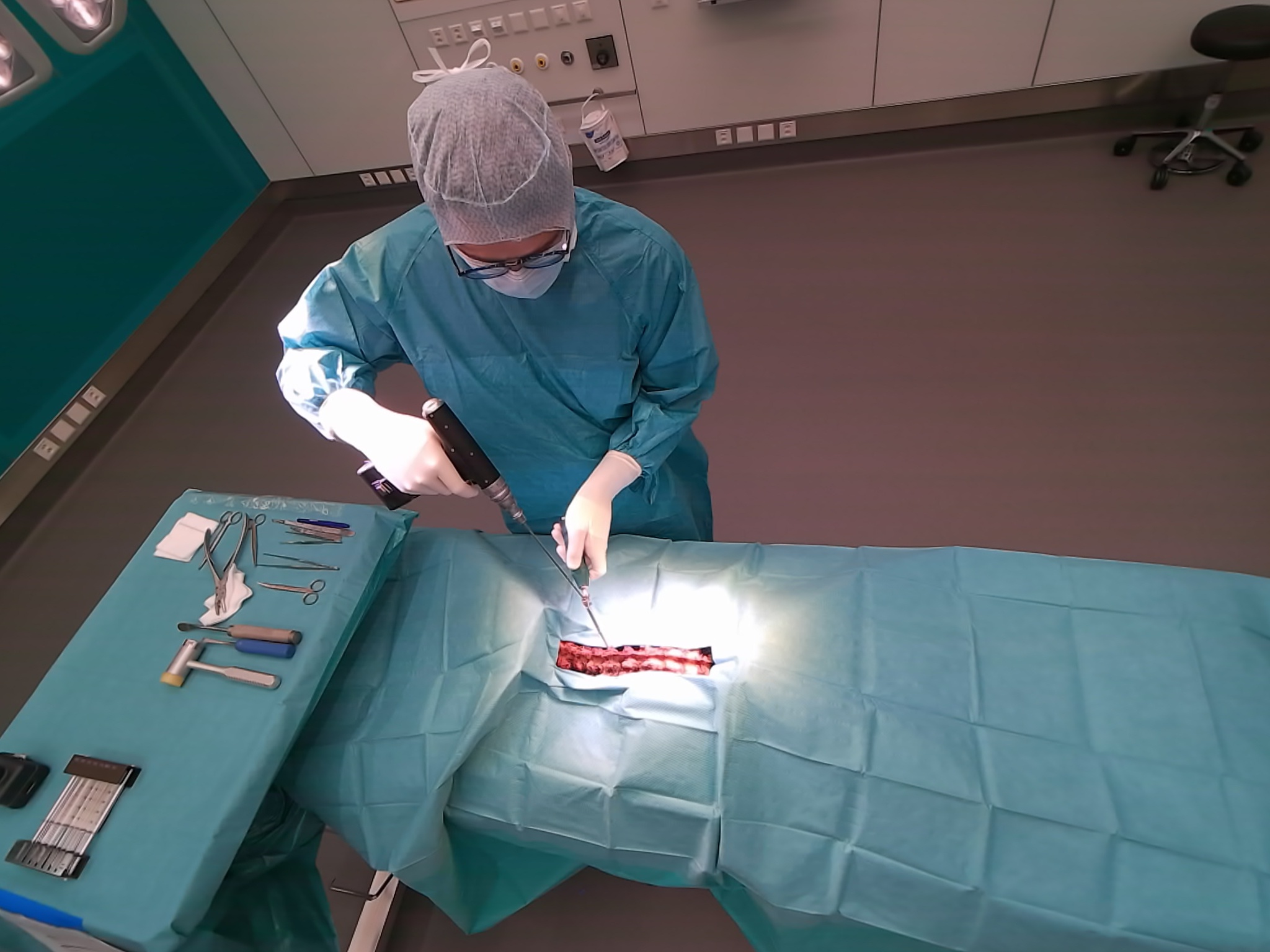}
    \includegraphics[width=\linewidth, keepaspectratio]{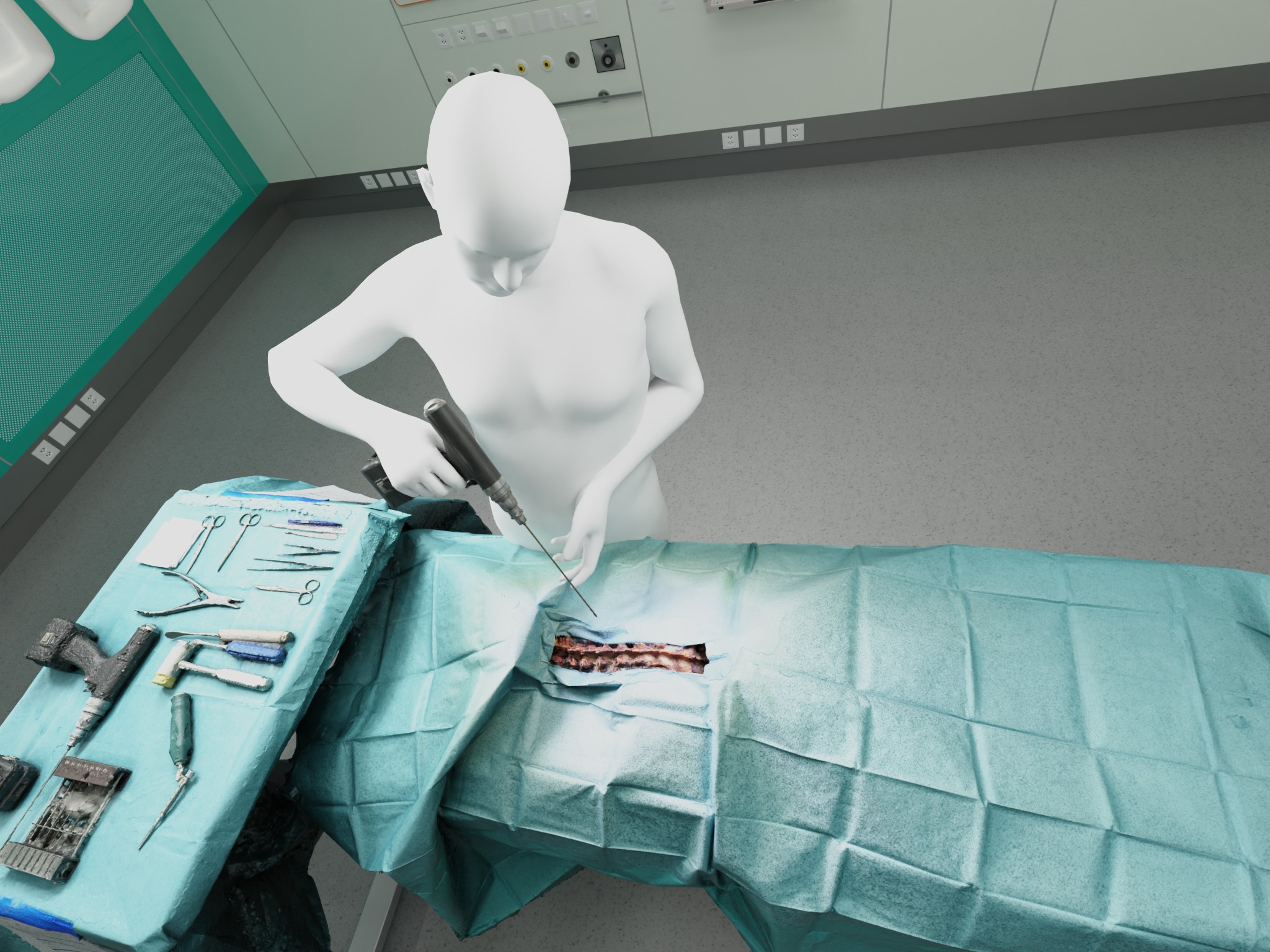}
    \end{minipage}%
    \hspace*{1mm}
    \begin{minipage}[b]{.195\linewidth-1mm}
    \adjincludegraphics[width=\linewidth, trim={{0.0\width} {.3\height} {0.0\width} {.1225\height}}, clip, keepaspectratio]{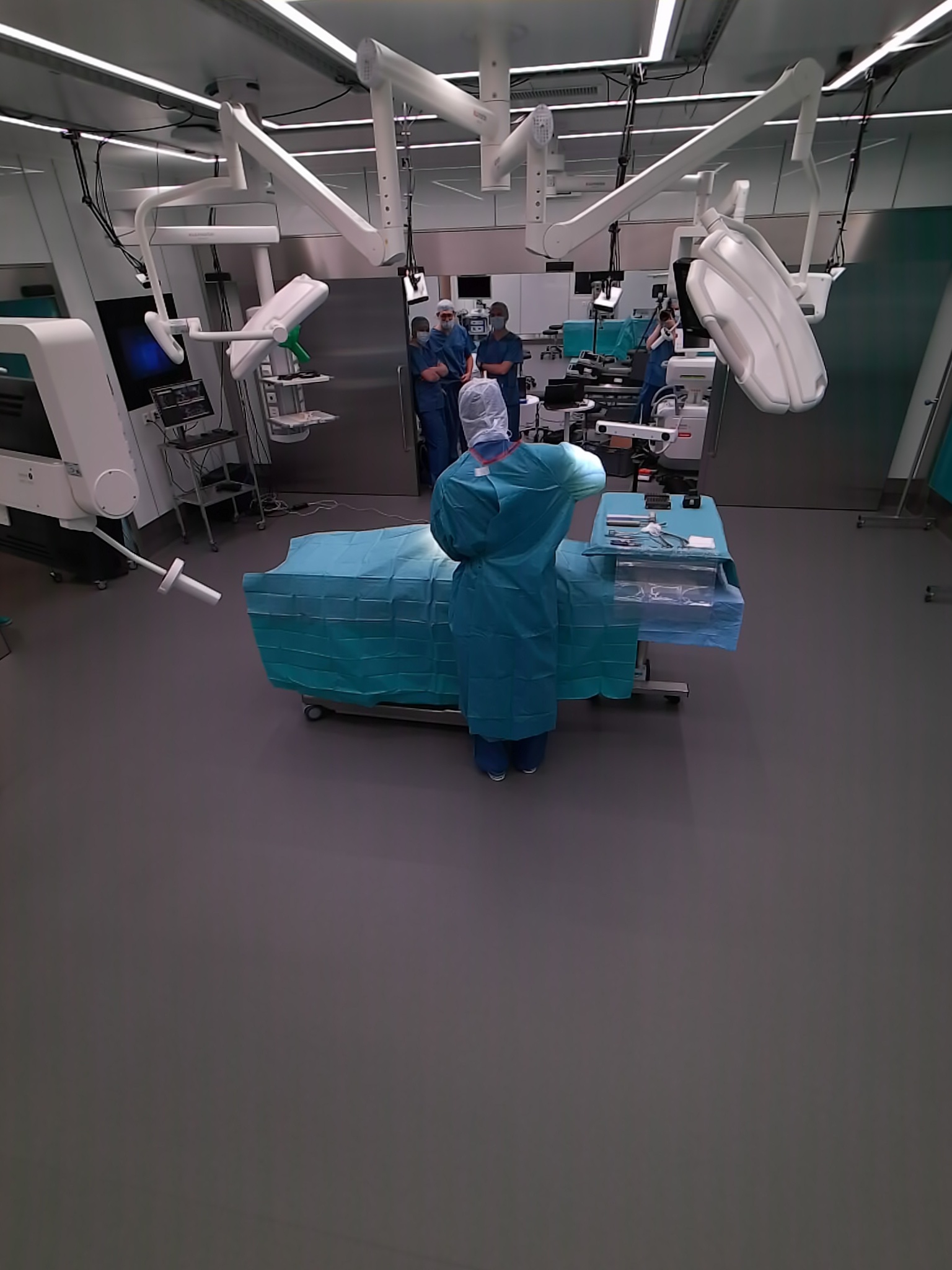}
    \adjincludegraphics[width=\linewidth, trim={{0.0\width} {.3\height} {0.0\width} {.1225\height}}, clip, keepaspectratio]{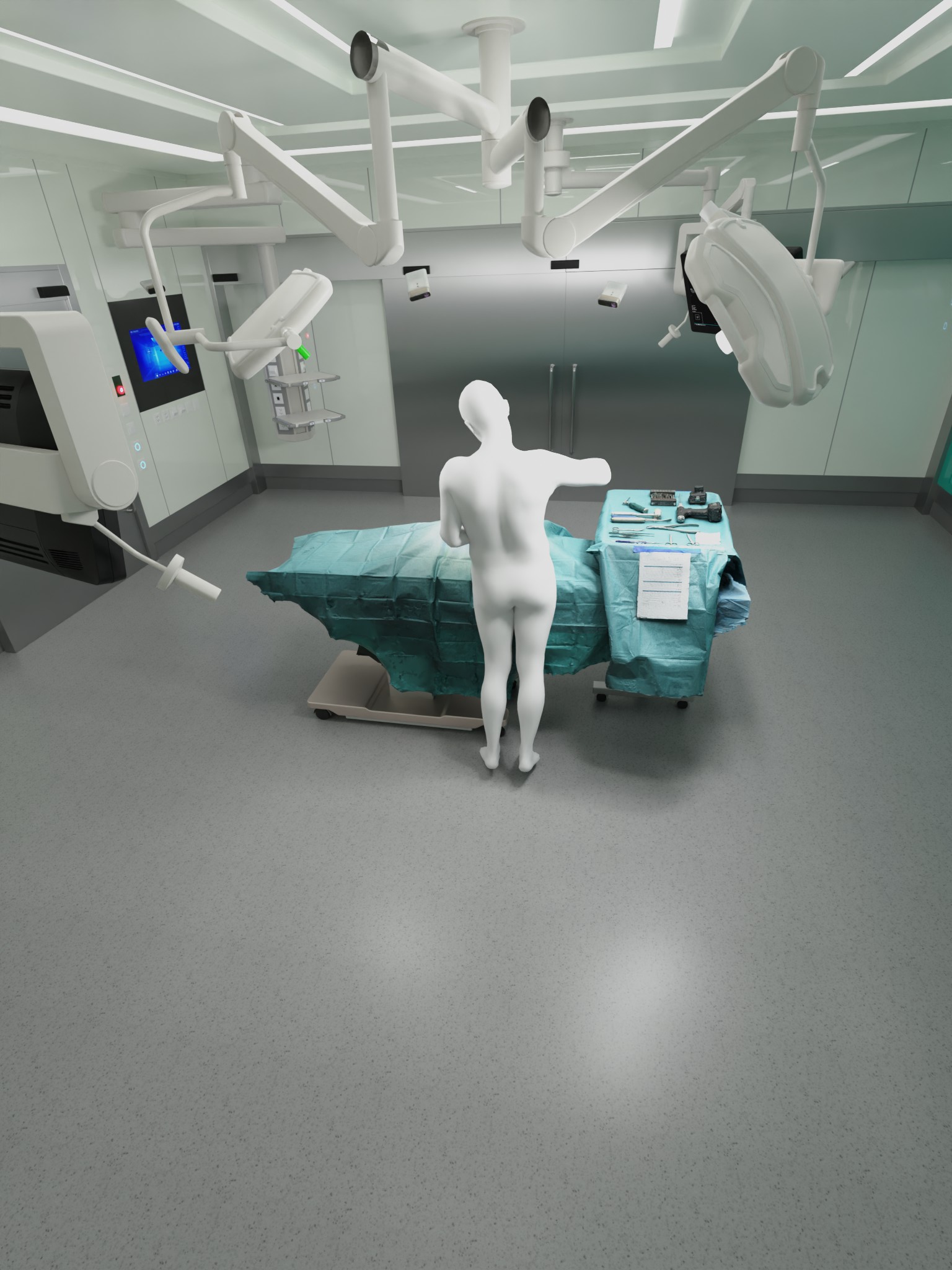}
    \end{minipage}%
    \caption{Comparison of the rendered digital twin with the real camera images. The camera perspectives shown from left to right correspond to the \textit{Kinect} cameras 1-5 as shown in Figure \ref{fig:camera_positions}. The digital twin was rendered in \textit{Blender} using the Cycles engine. 
    }
    \label{fig:synth_real_comparison}
\end{figure*}

To capture the dynamics of the scene, we deploy a motion capture setup comprising five ceiling-mounted \textit{Azure Kinect} RGB-D cameras (Microsoft Corporation, Redmond, WA, USA) and a \textit{FusionTrack 500} marker-based tracking system (Atracsys LLC, Puidoux, Switzerland).
We place four cameras opposite of the surgeon to capture their interaction with the instruments and patient, as shown in \mbox{Figure \ref{fig:camera_positions}}.
These cameras are mounted at different distances to the operating table, such that both the surgical near field and far field are captured.
As the surgeon's lower body is occluded by the operating table in all four cameras, we mount a fifth camera behind the surgeon to complement the four frontal viewpoints and simplify the body pose estimation task.
All cameras are mounted above the surgeon's head height to minimise the intrusiveness of our setup.
Due to weight limitations of the camera arms, the tracking system is placed on a tripod.
All RGB-D cameras are hardware-synchronized.
We follow the approach proposed by \cite{hein2023nextgeneration} to calibrate the cameras extrinsic parameters and to temporally synchronize them with the tracking system.
We then registered the multi-camera setup to the surgical environment by solving the \ac{pnp} problem between 3D points in the reference point cloud and corresponding 2D pixels in one of the cameras.
Hereby, we utilized the same 21 markers that were used to register the laser scans, as described in Section \ref{sec:reference_frame}.

\noindent\textbf{Surgical instruments}
Spinal instrumentation consists of pre-drilling screw trajectories for pedical screw placement.
Hereby, an AR-600 battery-powered drill (Arthrex Inc., Naples, FL, USA) is used along with a drill sleeve (Depuy Synthes, Raynham, MA, USA).
The drill is tracked via a marker array comprising five \ac{ir} reflective hemispheres with a diameter of \SI{3}{\milli\meter} attached to the drill body.
Following \cite{hein2023nextgeneration}, we obtain a 3D model of the surgical drill and marker array using a high-fidelity 3D scanner (Artec3D, Senningerberg, Luxembourg).
We registered the marker array to the 3D model by aligning virtual spheres to each hemisphere using the \ac{icp} algorithm.
We obtain a second 3D scan without any attached markers in order to hide the attached marker in the renderings.
Both 3D scans are registered using \ac{icp}.

\noindent\textbf{Human pose estimation}
To recover the surgeon's body pose, we fit the SMPL-H model \citep{SMPL:2015, MANO:SIGGRAPHASIA:2017} to the multi-view RGB images.
We detect 2D keypoints for all RGB images via \textit{OpenPose} \citep{openpose2019}. 
These keypoints describe 25 distinct anatomical locations of the human body as well as 21 locations on each hand. 
To take into account that multiple individuals would be present in a real surgery, we follow a simple heuristic to select the keypoints corresponding to the surgeon by computing the mean for each identified person and selecting the person closest to the operating table. 
Given the 2D keypoints detected in all images and the camera calibration, we compute the 3D keypoints of the surgeon via triangulation.
Lastly, we run a multi-stage optimization algorithm over the 2D and 3D keypoints, 2D bounding boxes, and SMPL-H model parameters and weights to compute the SMPL-H model. 
The optimization algorithm strongly resembles the one from \textit{EasyMocap}\footnote{\url{https://github.com/zju3dv/EasyMocap}}, but we added a moving average smoothness term to reduce jittering.

\section{Results}\label{sec:results}

While acknowledging the preliminary nature of this research and the requirement for manual intervention, we provide both quantitative and qualitative results to demonstrate the feasibility, accuracy, and potential benefits of our \ac{poc}.

\noindent\textbf{Quantitative results}
The registration error for the laser scan fusion is reported using the root mean square errors (RMSE) from point-to-point registrations performed between two laser scans, namely source and target ones.
One laser scan was used as target for all the registrations and all the other ones as sources. 
We additionally report the 3D reconstruction accuracy for the \ac{or} in terms of \ac{cd} over overlapping areas between two laser scans.
The per-registration RSME and \ac{cd} reported in Table \ref{tab:laser_scan_registration_error} verify that the fused reference point cloud is millimeter-accurate on the scale of the operating room. We furthermore evaluate the registration error between the laser scans and the photogrammetry model in terms of one-sided \ac{cd} from the fused laser scans to the photogrammetry model.
The obtained \ac{cd} is \SI{6.72}{\milli\meter}.
These results show that our \ac{sdt} is generally millimeter-accurate, an accuracy which may be partially attributed to the rigidity assumption that holds for our experimental setup.

Following \cite{hein2023nextgeneration}, we evaluate the calibration and temporal synchronization errors of the RGB-D cameras in terms of reprojection errors.
The errors are averaged over all frames in the calibration sequence and reported in Table \ref{tab:extrinsics_synchronization_errors}.
We additionally evaluate the registration of the camera array to the reference point cloud by computing the reprojection errors of the reference markers (as discussed in Section \ref{sec:reference_frame}) into each camera.
We obtain an average reprojection error of $1.39$ pixels.

\begin{table}[t]
\centering
\caption{Point-to-point registration errors of the laser scans. We choose the laser scan with most visible markers as the reference and register the remaining 7 scans based on all markers visible whose number is indicated in the first row. We report the RMSE of the registered 3D marker positions as well as \acf{cd} between both point clouds, with an outlier filtering of \SI{0.1}{\meter}.}
\label{tab:laser_scan_registration_error}
\resizebox{\columnwidth}{!}{%
\begin{tabular}{lcccccccc}
Laser Scan & 1    & 2    & 3    & 4    & 5    & 6    & 7    & Mean \\ \hline
\# Markers & 12 & 13 & 13 & 14 & 12 & 13 & 12 & 12.7 \\
RMSE (mm)  & 7.81 & 6.42 & 6.72 & 5.79 & 6.95 & 8.16 & 6.03 & 6.84 \\
CD (mm)    & 4.47 & 5.02 & 4.90 & 4.08 & 2.90 & 3.50 & 3.81 & 4.10
\end{tabular}%
}
\end{table}

\begin{figure*}[t]
    \centering
    \begin{minipage}[b]{.33\linewidth}
    \includegraphics[width=\linewidth, keepaspectratio]{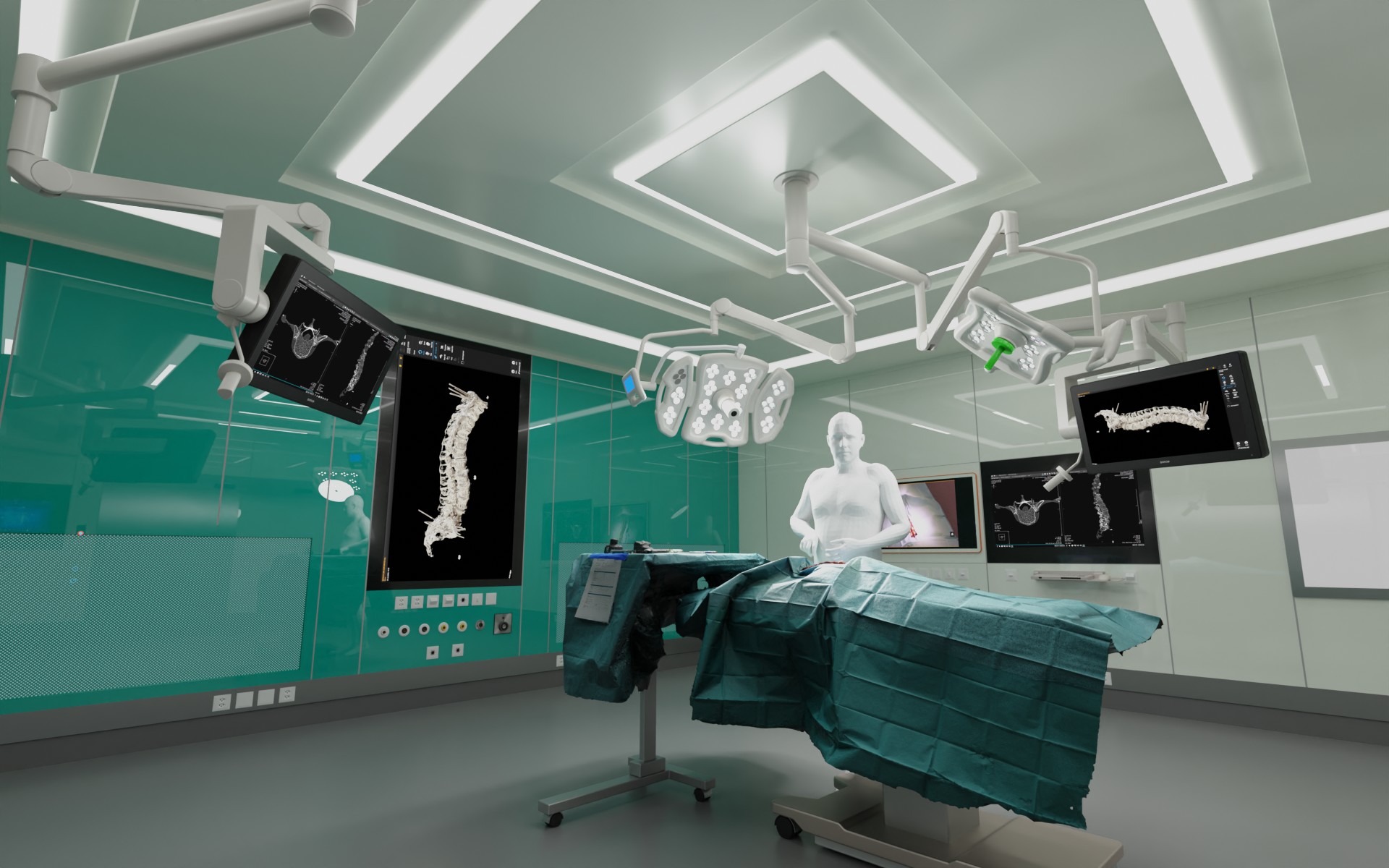}
    \includegraphics[width=\linewidth, keepaspectratio]{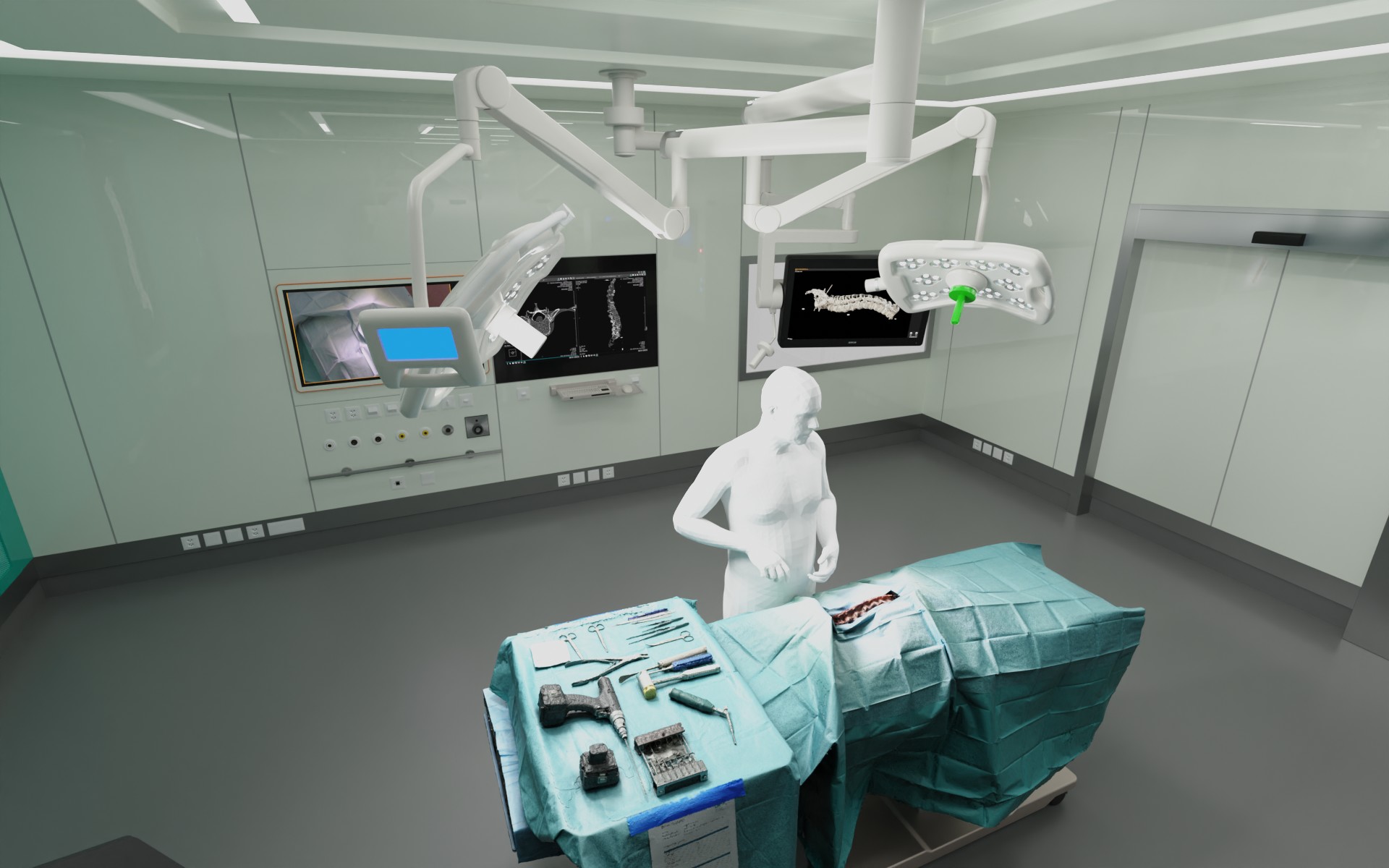}
    \end{minipage}%
    \hspace*{1mm}
    \begin{minipage}[b]{.33\linewidth}
    \includegraphics[width=\linewidth, keepaspectratio]{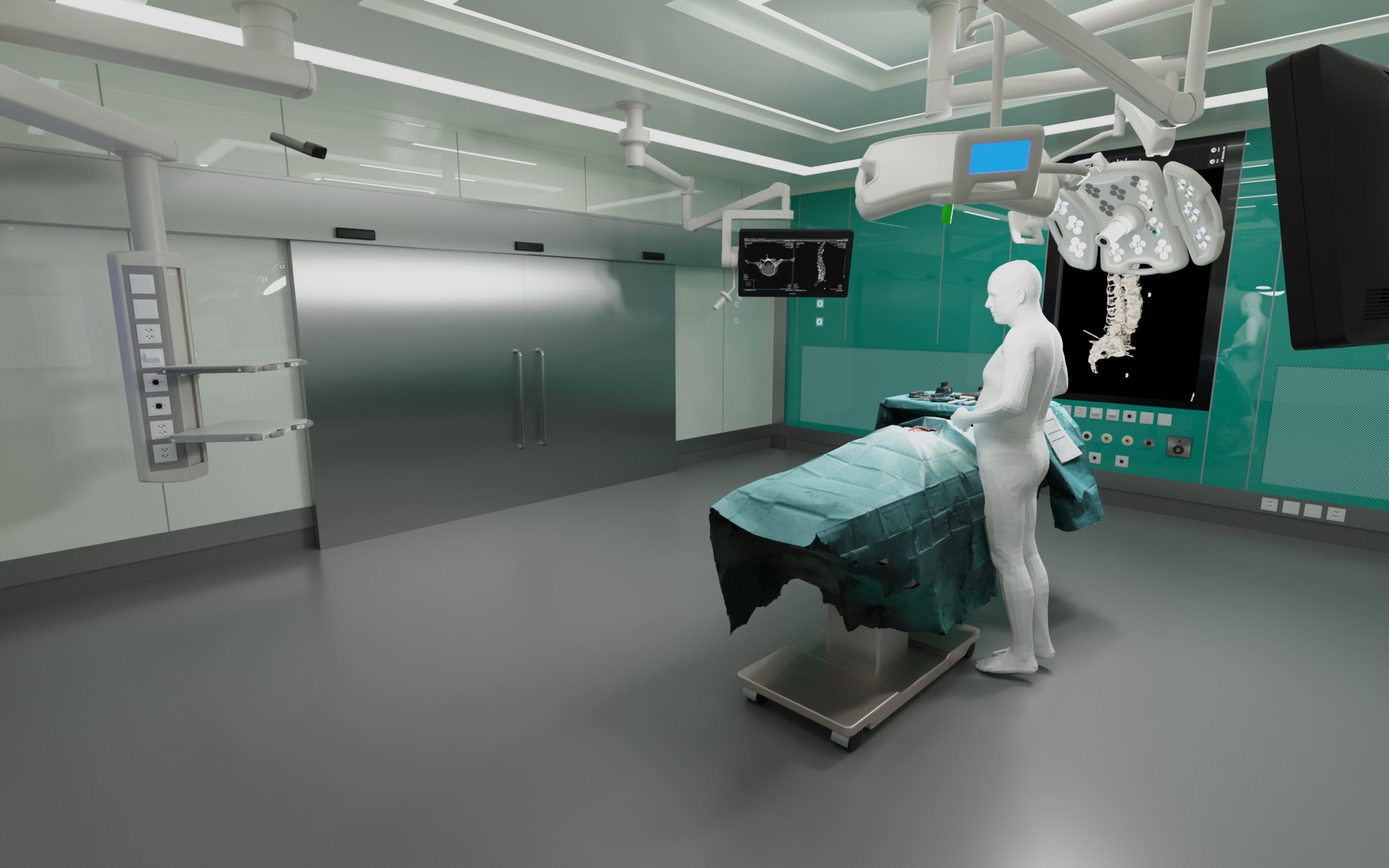}
    \includegraphics[width=\linewidth, keepaspectratio]{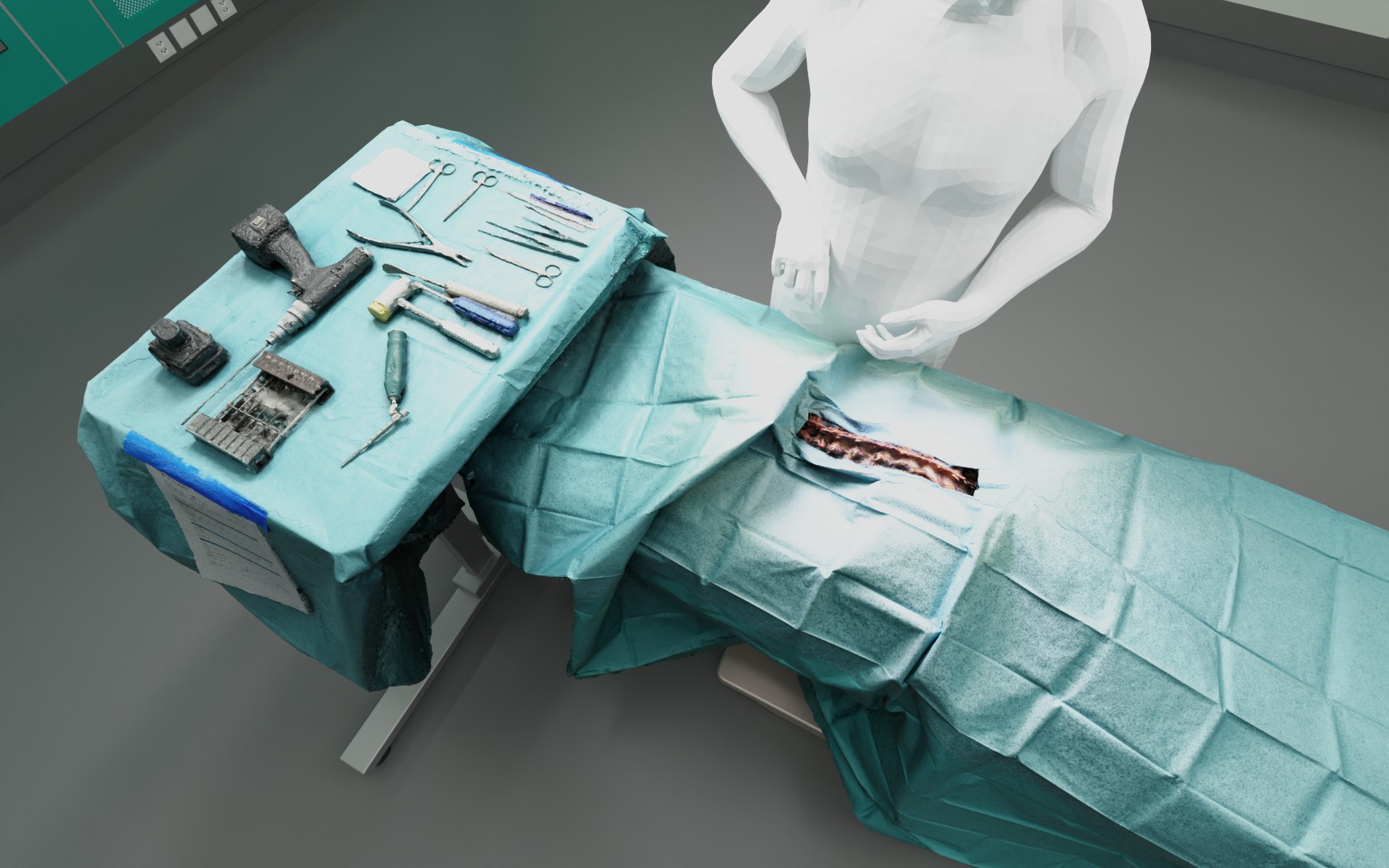}
    \end{minipage}%
    \hspace*{1mm}
    \begin{minipage}[b]{.33\linewidth}
    \includegraphics[width=\linewidth, keepaspectratio]{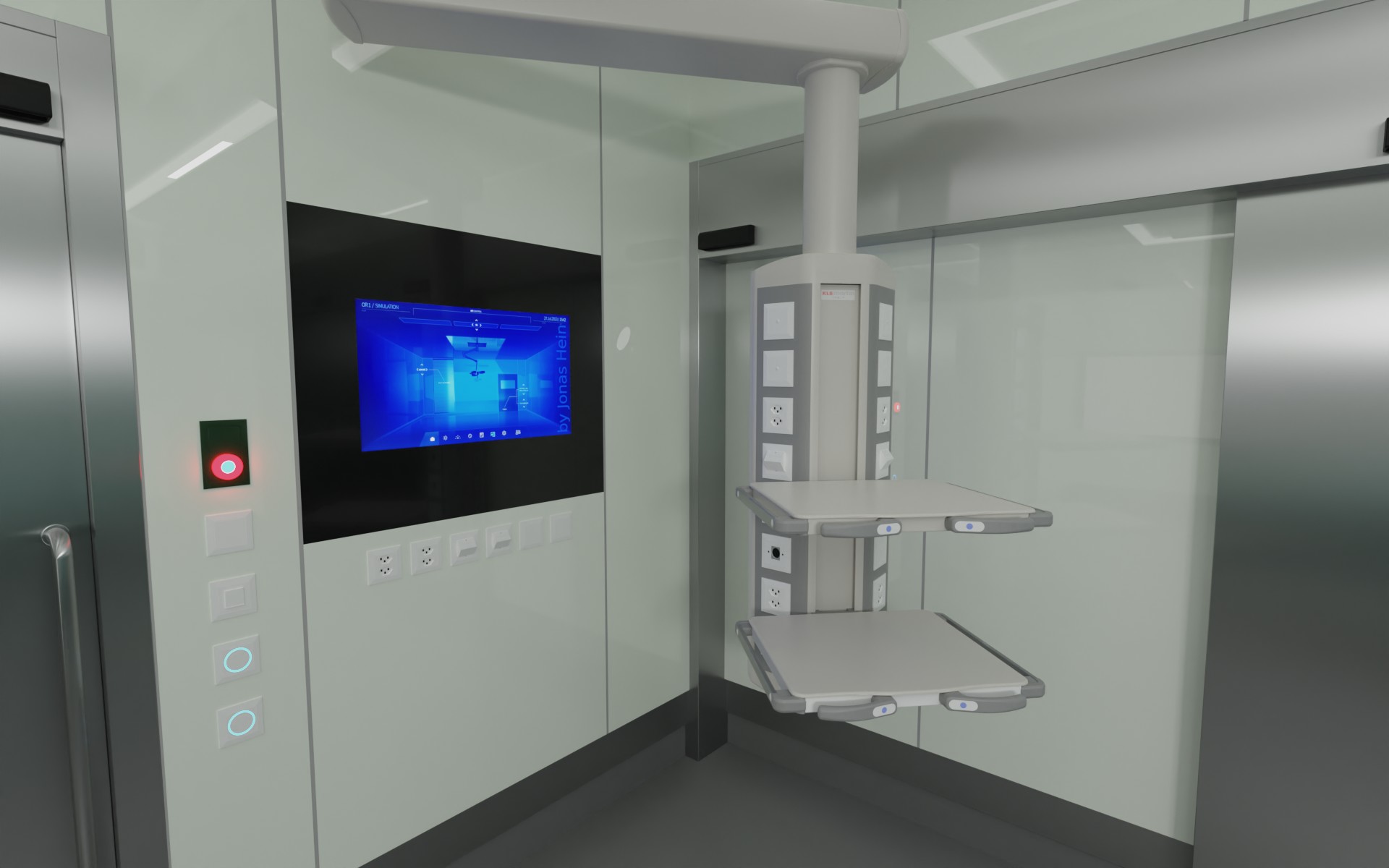}
    \includegraphics[width=\linewidth, keepaspectratio]{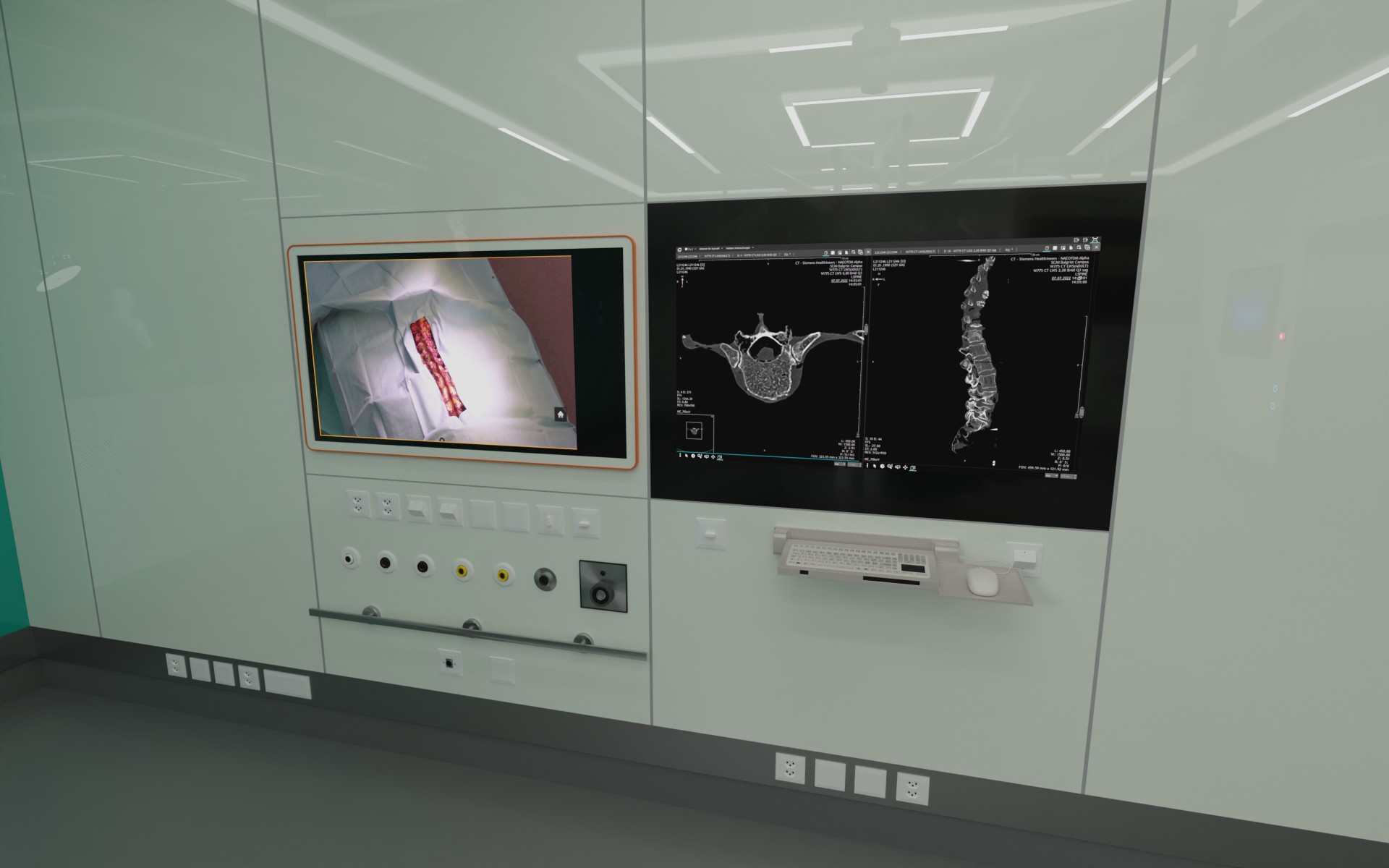}
    \end{minipage}
    \caption{Exemplary renderings of the operating room including the reconstructed operating table and the surgeon's estimated body pose.}
    \label{fig:operating_room_render}
\end{figure*}

\begin{table}[t]
\centering
\caption{Reprojection errors after the extrinsics calibration and synchronization of each RGB-D camera to the tracking system. 
We report the mean and standard deviation of reprojection errors over all frames in the calibration sequence. 
The camera locations are visualized in Figure \ref{fig:camera_positions}.}
\label{tab:extrinsics_synchronization_errors}
\resizebox{\columnwidth}{!}{%
\begin{tabular}{lcccccc}
Camera & 1 & 2 & 3 & 4 & 5 & Mean \\ \hline
Mean error (px)         & 0.75 & 0.40 & 1.06 & 1.63 & 0.39 & 1.19 \\
Std of errors (px) & 0.36 & 0.29 & 0.89 & 1.12 & 0.38 & 0.92 \\
\end{tabular}
}
\end{table}



\noindent\textbf{Qualitative results}
To showcase our digital twin, we render a video of the spatio-temporal scene with different visual overlays, which is available in the supplementary material. %
The video is rendered in \textit{Blender 3.3.1} using the Eevee rendering engine. 
Subtitles, transitions and side-by-side comparisons are added in post-processing using \textit{DaVinci Resolve} (Blackmagic Design Pty. Ltd, Port Melbourne, Australia).

The point clouds computed for each RGB-D camera are combined and outlier filtering is applied.
The filtered point cloud is cropped using manually defined bounding boxes to remove static elements like walls and floor. 
The cropped point cloud is voxelized to obtain a cleaner look with a uniform density.
Additional static rendered images are also shown in Figures \ref{fig:synth_real_comparison} and \ref{fig:operating_room_render}.

\section{Discussion}\label{sec:discussion}

In this section, we discuss the challenges we encountered that justify the sensors being employed and the proposed methodology. We also outline the main limitations of our \ac{poc} and discuss directions for future work.


\noindent\textbf{Challenges}
The generation of this digital twin highlighted several surgery-specific challenges.
Firstly, the frequent use of glass and metal surfaces in operating rooms poses a significant challenge for optical sensors and systems due to their reflectivity. 
We tested two commercial 3D scanners and a photogrammetry approach for the reconstruction of the operating room, but all failed to reconstruct the glass-covered walls or the metal operating tables. 
These early results also motivated us to use a laser scanner for the generation of a reference point cloud, which greatly simplified the registration of all entities in a common coordinate frame.

Secondly, the 3D reconstruction of human anatomy requires high-resolution imaging to capture its complex geometry and fine detailed texture.
The resolution of the \textit{Faro} laser, designed to capture large objects and environments, has shown to not be sufficient.
This motivates the use of photogrammetry to reconstruct the operating table and anatomy, a technology that shows a good compromise between acquisition time and reconstruction accuracy at this scene scale.

Thirdly, the surgical scrubs posed a challenge for human body pose estimation method and specifically for the keypoint detector, which were trained on humans wearing casual clothing.
Refining the pretrained models on medical staff wearing scrubs, e.g. using the MVOR dataset \citep{srivastav2018mvor}, could yield a domain-specific model with an improved performance.

\newpage

\noindent\textbf{Limitations}
Our current implementation has several limitations that need to be addressed.
Firstly, the photogrammetry-based reconstruction of the incision does not capture any dynamics, so the reconstruction of a full surgery would require multiple captures at different key steps as a minimum.
The time-consuming capture process makes this approach unfeasible for real surgery.
Instead, recent dynamic 3D reconstruction approaches based on NeRFs \citep{Mihajlovic:ResFields:2023} or Gaussian splatting \citep{wu20234dgaussians} could be utilized to reconstruct dynamic surfaces in the scene and specifically the incision.
The limited changes of the patient anatomy during the step of pedicle screw placement motivated us to rely on a photogrammetry approach for this prototype.

Secondly, our proof-of-concept assumes rigidity of the instrument table, operating table, and \ac{or} lamps and displays. 
Articulations and movements of the operating tables could instead be tracked by continuously registering the 3D model to the dynamic point cloud, via \ac{ml}-based pose estimation methods \citep{li2020category}, or by utilizing marker-based tracking.
In a similar fashion, a CT-based 3D model of the patient anatomy could be registered to the spatio-temporal scene by either point cloud-based registration \citep{liebmann2024automatic} or via marker-based tracking. 
Also, the static display contents in our \ac{poc} could be replaced by a screen recording or a lightweight state-based representation of the shown contents.

Thirdly, the available data streams are added independently to our digital twin. 
As a result, calibration errors and noise can cause inconsistencies in the shared spatio-temporal representation, e.g. a mismatch in the hand and instrument pose. 
These inconsistencies could be reduced by integrating sensors jointly or based on a learnt model, such that the spatio-temporal consistency and plausibility of the digital twin can be enforced.

Finally, the generation of this prototype was time-consuming due to the lack of automated processes. 
Several steps of our pipeline, such as the registration of overhead devices, the capture of close-range photographs for the photogrametric reconstruction of the operating table, or the registration of the anatomic model were conducted by hand. 
To enable a systematic and efficient generation of \ac{sdt}, these manual steps need to be automated.


\noindent\textbf{Future work}
Evolving our presented \ac{poc} into a complete \ac{sdt} requires the integration of further sensors, an automated interpretation to generate semantic labels, and ideally the inclusion of prior knowledge.
First, the integration of additional sensors like microphones, patient vitals, and medical imaging ensure that the digital twin accurately captures the available information at a time.
Naturally, the most relevant sensors to monitor the patient already exist in today's \ac{or}, which reduces the cost of integrating additional sensors.
Second, the data streams from all sensors need to be automatically analysed and interpreted to extract semantics.
The method of estimating the surgeon's body pose is representative for further extensions of our digital twin with estimated semantic annotations, for example from segmentation \citep{ma2024segment}, surgical scene graphs \citep{lin2023sgt++}, anatomical landmarks \citep{zhu2021you}, or surgical phase detection \citep{czempiel2021opera}.
Last, the integration of prior knowledge in form of physical, \mbox{(bio-)mechanical} \citep{dao2017multimodal}, or behavioral models \citep{Zhao23} is needed to extend the presented spatio-temporal reconstruction to a comprehensive digital twin.

\section{Conclusion}\label{sec:conclusion}
In this work, we presented a proof-of-concept for surgery digitalization. 
We outlined the potential of \ac{sdt}s for education, training data generation, simulation and closed-loop optimization, and automation of surgical tasks such as planning and reporting. 
We proposed a methodology to obtain an \ac{sdt} encompassing the most relevant entities in surgery.

In contrast to related works that manually craft a virtual environment to simulate surgeries, our approach focuses on the capture of a real surgery. 
In its current state, our prototype can already be used to capture and re-render surgical steps or simple interventions for educational purposes, for example in the form of training videos or interactive \ac{vr}-based applications.

Our PoC is a step towards the systematic capture of surgeries, which may be used to collect a large dataset of digitized surgeries, including rare pathological cases and other infrequent events such as unforeseen complications or surgical errors. 
Moreover, the generated \ac{sdt}s can provide a realistic environment for the training of \ac{ml}-based models and robotic agents with a reduced sim-to-real gap.
In the long run, holistic approaches to surgery digitization may boost the performance of state-of-the-art methods in computer-assisted surgery due to comprehensive representations of the current state of the surgery.
We hope that our work motivates further research on automated methods for surgery digitization and the creation of \ac{sdt}s.


\section*{Data availability}
The data will be made available on our project page \resizebox{\linewidth}{!}{\url{https://jonashein.github.io/surgerydigitization/}.}

\section*{Acknowledgements}
This work has been supported by the OR-X - a swiss national research infrastructure for translational surgery - and associated funding by the University of Zurich and University Hospital Balgrist, as well as by the InnoSuisse PROFICIENCY grant.
The study was conducted according to the guidelines of the Declaration of Helsinki, and approved by the local ethical committee (KEK Zurich BASEC No. 2021-01196). 
We thank Jan David Grunder for his support in the creation of the CAD model of the OR.

{
    \small
    \bibliographystyle{ieeenat_fullname}
    \bibliography{main}
}


\end{document}